\documentclass[10pt,twocolumn,letterpaper]{article}

\usepackage[pagenumbers]{cvpr} 

\definecolor{cvprblue}{rgb}{0.21,0.49,0.74}
\usepackage[pagebackref,breaklinks,colorlinks,allcolors=cvprblue]{hyperref}

\usepackage{times}
\usepackage{epsfig}
\usepackage{graphicx}
\usepackage{amsmath,amssymb,amsthm}
\usepackage{booktabs}
\usepackage{adjustbox} 
\usepackage{array} 
\usepackage{multirow}
\usepackage{mathtools}
\usepackage{algorithm}
\usepackage{algorithmic}
\usepackage{bm}
\usepackage{enumitem}
\usepackage{color}
\usepackage{wrapfig}
\usepackage[accsupp]{axessibility}
\usepackage{microtype}
\usepackage{dsfont}
\usepackage{pifont}
\usepackage{hyperref}
\hypersetup{colorlinks,linkcolor=blue,citecolor=blue,urlcolor=blue}
\usepackage[table]{xcolor}
\usepackage{colortbl}
\usepackage{comment}

\def\paperID{4599} 
\def\confName{CVPR}
\def\confYear{2026}

\title{EatGAN: An Edge-Attention Guided Generative Adversarial Network for Single Image Super-Resolution}

\author{
Penghao Rao \qquad Tieyong Zeng\thanks{Corresponding author.}\\
Department of Mathematics, The Chinese University of Hong Kong\\
{\tt\small basillowe@link.cuhk.edu.hk \quad zeng@math.cuhk.edu.hk}
}

\begin{document}
\maketitle
\begin{abstract}
\label{sec:abstract}
Single-image super-resolution (SISR) is an important task in image processing, aiming to enhance the resolution of imaging systems. Recently, SISR has made a significant leap and achieved promising results with deep learning. GAN-based models stand out among all the deep learning models because of their excellent performance in perceiving quality. However, it is rather difficult for them to reconstruct realistic high-frequency details and achieve stable training. To solve these issues, we introduce an Edge-Attention guided Generative Adversarial Network (EatGAN), the first GAN-based SISR model that simultaneously leverages edge priors both explicitly and implicitly inside the generator, which (i) proposes a Normalized Edge Attention (NEA) mechanism based on channel-affine and spatial gating that transforms edge prior into lightweight, learnable modulation parameters and injects and fuses them multiple times in a (ii) edge-guided hybrid residual block, which progressively enforces structural consistency across scales; and (iii) a composite generator objective combining pixel, perceptual, edge-gradient, and adversarial terms. Experiments show consistent state-of-the-art across distortion-oriented benchmarks and perception-oriented benchmarks. Notably, our model achieves 40.87 dB and 0.073 (LPIPS) on Manga 109, which indicates that reframing image priors from passive guidance into a controllable modulation primitive for generators can chart a practical path toward trustworthy, high-fidelity Super-Resolution.
\end{abstract}    
\section{Introduction}
\begin{figure}[t]
    \centering
    \includegraphics[width=0.45\textwidth]{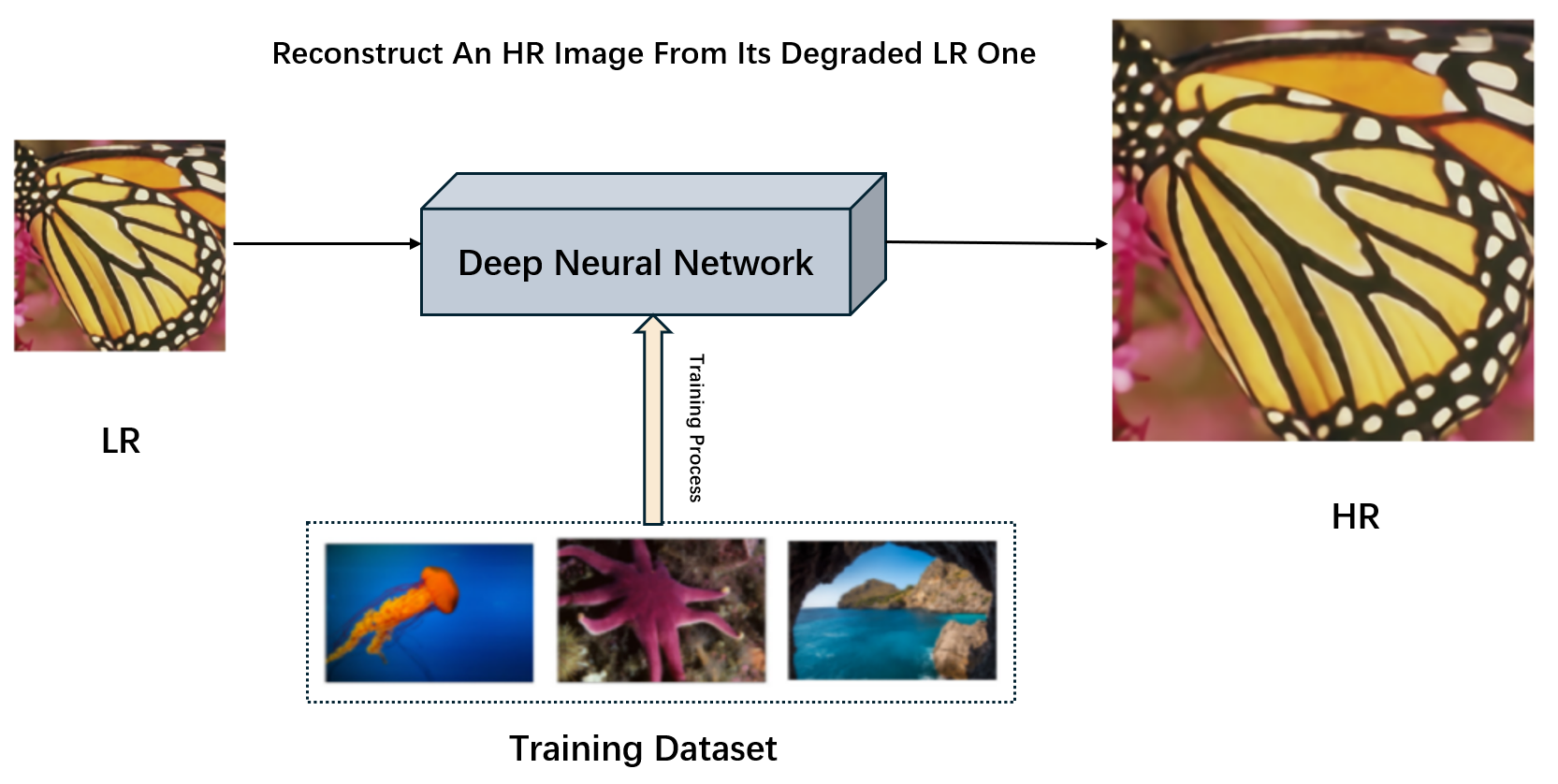}
    \caption{Given a single LR image, A deep neural network $f_\theta$, which has been well trained on a dataset to learn the mapping from LR images to their corresponding HR versions, generates its HR reconstruction with enhanced visual quality and sharper details.}
    \label{fig:workflow}
\end{figure}
Over the past few decades, image super-resolution, especially single-image super-resolution (\textbf{SISR}), has been extremely popular in the field of computer vision, aiming to reconstruct a super-resolution (\textbf{SR}) image from a single low-resolution (\textbf{LR}) image. It has various applications including medical image enhancement \cite{hyun2018deep}, \cite{ravishankar2019image}, video super-resolution \cite{ben2005video}, \cite{belekos2010maximum}, and facial illusion \cite{baker2000hallucinating}, \cite{baker2002limits}. Many SISR methods, based on interpolation \cite{duchon1979lanczos}, have been studied long before, but SISR is an inherently ill-posed problem, and multiple high-resolution (\textbf{HR}) images always correspond to the same LR image \cite{li2024systematic}. Hence, some numerical methods \cite{sun2008image,kim2010single} utilizing prior information and learning-based methods \cite{chang2004super,yang2010image} are proposed to address this problem. With the rapid development of deep learning (\textbf{DL}) techniques, shown in the \cref{fig:workflow}, numerous DL-based methods have been proposed for SISR, continuously showing State-Of-The-Art (\textbf{SOTA}). Therefore, how to construct a concise and efficient model to finish this task becomes a heated discussion topic. It is well-known that DL-based methods can be divided into supervised and unsupervised methods. For supervised learning methods, the LR and HR images have a one-to-one correspondence, and researchers compute the reconstruction error between the ground-truth image and the reconstructed image, or search for a mapping to transform the image maps to another space and then minimize the distance between the reconstructed image and the ground-truth image. However, the real paired images are difficult to collect. Hence, the unsupervised learning method becomes widely used. This type of method \cite{shocher2018zero,yuan2018unsupervised,zhu2017unpaired} no longer uses paired LR-HR images for training but uses unpaired LR-HR images or itself. Among them, models based on Generative Adversarial Networks (\textbf{GAN}) have outstanding advantages, since GAN \cite{goodfellow2014generative} can make the reconstructed SR image more realistic. For example, Ledig et al. \cite{ledig2017photo} form the Super-Resolution Generative Adversarial Network (SRGAN), which has been widely used in the single-image super-resolution field nowadays. Wang et al. \cite{wang2018esrgan} make modifications to SRGAN and propose ESRGAN, which improves the generalization ability. In SRFeat \cite{park2018srfeat}, Park et al. indicate that the GAN-based SISR methods tend to produce less meaningful high-frequency noise in reconstructed images, so they adopt the image discriminator. Moreover, another team \cite{wang2023high} proposes a novel GAN framework that utilizes the powerful generative ability of StyleGAN-XL. However, all these methods have an intractable drawback: unstable training.

Although current SISR models have made significant breakthroughs, how to leverage information inside and outside the image to further improve model performance remains worth exploring. One of these methods is prior-guided SISR frameworks. For example, SFTGAN \cite{wang2018recovering} uses the semantic categorical prior to generate richer and more realistic textures, SPSR \cite{ma2020structure} utilizes the gradient maps to guide image recovery, and FeMaSR \cite{chen2022real} uses discrete features obtained by VQ-GAN \cite{yu2021vector} as prior information to perform image recovery. Among all these image prior methods, edge prior often achieves SOTA. Yang et al. \cite{yang2017deep} integrated the edge prior with recursive networks and proposed a Deep Edge Guided Recurrent Residual (DEGREE) Network. After that, Fang et al. \cite{fang2020soft} proposed an efficient and accurate Soft-edge Assisted Network (SeaNet). However, they are all CNN-based methods. Currently, there is no GAN-based model that can effectively utilize image edge priors in its generator.

To improve the efficiency during the learning procedure, attention mechanism has been proposed. Former works \cite{hu2018squeeze,wang2018non} use it to guide the network to pay more attention to the regions of interest. Motivated by these methods, the attention mechanism has also been introduced into SISR. Many methods \cite{mei2018effective,zhang2018image,dai2019second} introduce the SE mechanism in the SISR model. When CNN-based methods conduct convolution in a local receptive field, the contextual information outside this field is ignored, while the features in distant regions may have a high correlation and can provide effective information. Given this issue, non-local attention
has been proposed as a filtering algorithm to compute a weighted mean of all pixels of an image. Multiple experiments show that these types of methods \cite{liu2018non,zhang2019residual,niu2020single,mei2020image,xia2022efficient}, which use non-local attention, can further improve the model performance. However, existing attention mechanisms still suffer from problems such as inaccurate structural localization.

\begin{figure*}[t]
    \centering
    \includegraphics[width=1.0\textwidth]{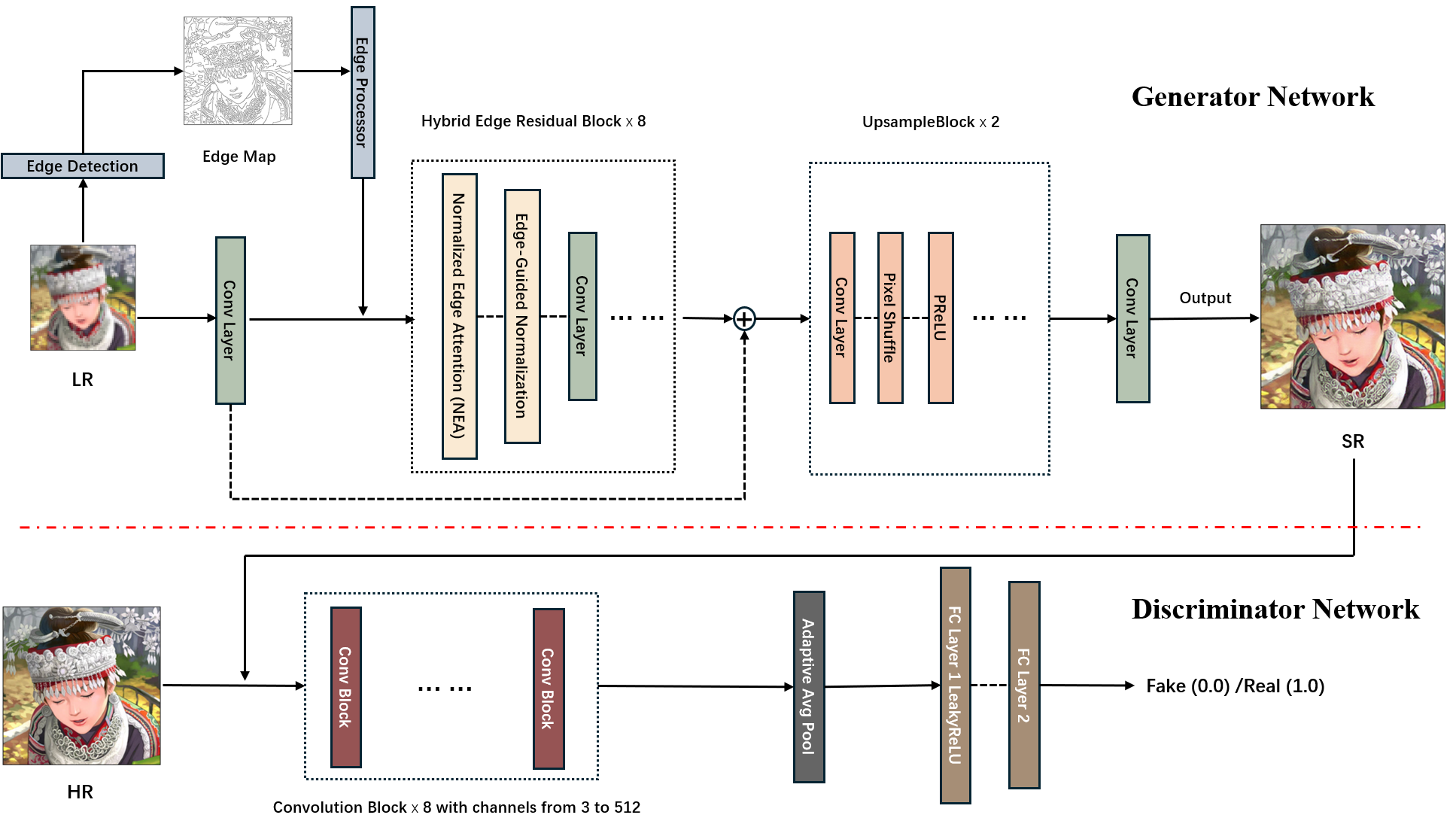}
    \caption{Overview of EatGAN Architecture. \textbf{Top:} The generator takes LR images and extracted edge maps as inputs, processes them through hybrid edge residual blocks with Normalized Edge Attention (NEA) mechanisms, and generates SR images. \textbf{Bottom:} The discriminator consists of 8 convolution blocks, global pooling, and fully connected layers for fake (0.0) and real (1.0) classification.}
    \label{fig:EatGAN_architecture}
\end{figure*}

To address these issues, we propose the Edge-Attention guided Generative Adversarial Network (EatGAN), the first GAN-based model to implicitly and explicitly use image edge information simultaneously in the generator. In this model, we utilize a composite loss function to solve the unstable training problem and use a normalized edge attention mechanism, focused on structurally significant high-gradient regions. The architecture of our model is shown in \cref{fig:EatGAN_architecture}.

The main contributions of this paper are:
\begin{itemize}
\item \textbf{Normalized Edge Attention Mechanism (NEA):}
We propose NEA which combines channel-affine modulation and spatial gating under unified normalization, enabling the network to focus on structurally significant regions while suppressing spurious high-frequency artifacts.

\item \textbf{Edge-Generator Loss and Hybrid Edge Residual Block:}
We introduce an edge gradient loss integrated with the standard objective to form a new generator loss. Meanwhile, we design a hybrid edge residual block that implicitly leverages edge information for image restoration.

\item \textbf{Empirical Results:}
Extensive experiments across both distortion- and perception-oriented benchmarks show that our model can achieve superior performance compared to SOTA methods, demonstrating that learning edge prior both implicitly and explicitly in the generator offers a principled path toward distortion and perception trade-off.
\end{itemize}

\section{Related Work}
\label{sec:related}

\subsection{Image Super-Resolution}
Image SR is a classic technique to improve the resolution of an imaging system, which can be classified into single-image super-resolution (SISR) and multi-image super-resolution (MISR) according to the number of input LR images. SISR is much more challenging since MISR has extra information for reference, while SISR only has information of a single input image\cite{li2024systematic}. With the rise of deep learning, DL-based techniques \cite{dong2015image,wang2020deep} have gradually become mainstream in the SR task. One prevalent approach of early works \cite{ahn2018image, alpher2002frobnication, kim2016accurate, wang2015deep} is to train a regression model using paired training data. To improve the perceptual quality of the reconstructed HR images, generative models \cite{dahl2017pixel, menick2018generating, den2016conditional, khan2022transformers} emerge and obtain significant improvements, but the computational cost also increases at the same time. Nowadays, diffusion-based models \cite{choi2021ilvr, chung2022come, guo2024make, kawar2022denoising, rombach2022high, saharia2022image} have been widely used in SR task. While promising results are achieved, they rely on a large number of inference steps, which greatly hinders their application. Besides, GAN-based methods \cite{guo2022lar, karras2017progressive, ledig2017photo, menon2020pulse, sajjadi2017enhancenet} are excellent in terms of perceptual quality, but the training is usually unstable. In this paper, we propose a GAN-based model with stable training to solve SISR task.

\subsection{Prior Guidence}
In the image recovery procedure, reconstructing realistic high-frequency details is difficult since many useful features have been lost or damaged. Hence, scientists propose the priors-guided framework. With the help of prior information, models can converge faster and achieve better reconstruction accuracy. There are multiple types of image priors, such as spatial prior \cite{wang2018recovering},  gradient maps prior \cite{ma2020structure}, discrete features prior \cite{chen2022real, yu2021vector}, edge prior \cite{yang2017deep, fang2020soft}, and so on. However, all these methods use it implicitly in the network, leading to a poor utilization rate. Fang et al. \cite{fang2020soft} have shown that more adequate utilization can lead to more significant performance. Therefore, in our model, we propose an edge-prior framework and modify the GAN generator to implicitly and explicitly utilize edge information simultaneously.

\subsection{Loss Function}
In the SISR task, the loss function is used to guide the iterative optimization process of the model by computing a certain kind of error. Researchers find that combining multiple loss functions can better reflect the situation of image restoration. Pixel loss \cite{elharrouss2025loss} aims to measure the difference between two images on a pixel basis, including the L1 loss, Mean Square Error (MSE) Loss, and Charbonnier loss. It is widely used but oversmooths details, weakens textures, and reduces perceptual fidelity. Content loss \cite{simonyan2014very, ledig2017photo} aims to measure the semantic difference between images. It is expressed as the Euclidean distance between the high-level representations of these two images. While it preserves semantic structures and improves perceptual realism, it may misalign colors. Adversarial loss \cite{goodfellow2014generative, ledig2017photo}, which aims to make the reconstructed SR image more realistic, consists of generator loss and discriminator loss. This produces realistic textures and sharpens details, but their hyperparameters are sensitive. Prior loss \cite{li2024systematic}, such as sparse prior loss \cite{wang2018recovering}, gradient prior loss \cite{ma2020structure}, and edge prior loss \cite{fang2020soft, yang2017deep}, is used depending on the type of prior information. Since these loss functions vary a lot, the results are also varied. Fourier space loss \cite{fuoli2021fourier} focuses on the frequency domain. It emphasizes high-frequency parts in the images and restores textures, but may cause ringing artifacts. To address various issues caused by these former loss functions, we design a new generator loss in GAN to enhance the generated images quality.

\subsection{Attention Mechanism}
Attention mechanism is a tool that can allocate available resources to the most informative part of the input. To improve the efficiency during the learning procedure, some works \cite{hu2018squeeze, wang2018non} are proposed to guide the network to pay more attention to the regions of interest. Channel attention \cite{zhang2018image, dai2019second} is proposed for improving flexibility in dealing with different types of information when using CNN-based models. This achieves remarkable results in image recovery, but there is a disadvantage that the features in distant regions, which may have a high correlation and can provide effective information, are ignored. Therefore, non-local attention \cite{liu2018non, zhang2019residual, niu2020single, mei2020image, xia2022efficient} has been proposed as a filtering algorithm to compute a weighted mean of all pixels of an image. This efficiently makes distant pixels contribute to the response of a position and further improves the model performance. But it increases the computation cost rapidly. Since all these attention mechanisms have their own disadvantages, to address these issues, we propose the Normalized Edge Attention (NEA) to improve our model's performance.

\section{Methodology}
\label{sec:method}
We propose an Edge-Attention guided GAN for single image super-resolution, which leverages edge prior, to enhance restoration quality. Our model consists of three key components: (i) A \textbf{Normalized Edge Attention (NEA)} mechanism that fuses edge guidance through spatial attention and edge-conditioned normalization pathways; (ii) \textbf{Hybrid Edge Residual Block} that progressively refine features with explicit structural constraints at multiple scales; and propose (iii) \textbf{A Edge Gradient Loss} combined with standard generator objective to improve generated image quality.
\begin{algorithm}[t]
\caption{EatGAN Training}
\label{alg:training}
\begin{algorithmic}[1]
\REQUIRE Pre-trained VGG-19 model $\phi(\cdot)$
\REQUIRE Paired training set $(X_{LR}, Y_{HR})$
\REQUIRE Edge detector $\mathcal{E}(\cdot)$ (Canny)
\STATE Initialize generator $G_\theta$ and discriminator $D_\phi$ from pre-trained weights
\WHILE{not converged}
    \STATE Sample mini-batch $(x_{lr}, y_{hr}) \sim (X_{LR}, Y_{HR})$
    \STATE Extract edge map: $e \leftarrow \mathcal{E}(x_{lr})$
    \STATE \textbf{Train Generator}
    \STATE Generate SR image: $\hat{y}_{sr} \leftarrow G_\theta(x_{lr}, e)$
    \STATE Compute pixel loss: $\mathcal{L}_{pixel} = \|\hat{y}_{sr} - y_{hr}\|^2$
    \STATE Compute perceptual loss: $\mathcal{L}_{perc} = \|\phi(\hat{y}_{sr}) - \phi(y_{hr})\|^2$
    \STATE Compute edge gradient loss: $\mathcal{L}_{edge} = \|\nabla\hat{y}_{sr} - \nabla y_{hr}\|^2$
    \STATE Compute adversarial loss: $\mathcal{L}_{adv}^G = -\log D_\phi(\hat{y}_{sr})$
    \STATE $\mathcal{L}_G = \mathcal{L}_{pixel} + \lambda_{perc} \mathcal{L}_{perc} + \lambda_{edge} \mathcal{L}_{edge} + \lambda_{adv} \mathcal{L}_{adv}^G$
    \STATE Update generator: $\theta \leftarrow \theta - \alpha \nabla_\theta \mathcal{L}_G$
    \STATE \textbf{Train Discriminator}
    \STATE Compute real loss: $\mathcal{L}_{real} = -\log D_\phi(y_{hr})$
    \STATE Compute fake loss: $\mathcal{L}_{fake} = -\log(1 - D_\phi(\text{detach}(\hat{y}_{sr})))$
    \STATE $\mathcal{L}_D = (\mathcal{L}_{real} + \mathcal{L}_{fake}) / 2$
    \STATE Update discriminator: $\phi \leftarrow \phi - \beta \nabla_\phi \mathcal{L}_D$
\ENDWHILE
\RETURN Trained generator $G_\theta$ and discriminator $D_\theta$
\end{algorithmic}
\end{algorithm}

\begin{table*}[htbp]
\centering
\caption{Quantitative comparison (average PSNR/SSIM) with other methods on standard benchmarks. Best results per column are highlighted. Methods are grouped by architecture type: \textbf{CNN-based}, \textbf{Transformer-based}, \textbf{Lightweight}, \textbf{Attention Mechanisms}.}
\begin{adjustbox}{width=1.0\textwidth}
\begin{tabular}{l|c|c|c|cc|cc|cc|cc|cc}
\toprule
Method & Scale & \#Params & \#FLOPs & \multicolumn{2}{c|}{Set5 \cite{bevilacqua2012low}} & \multicolumn{2}{c|}{Set14 \cite{zeyde2010single}} & \multicolumn{2}{c|}{BSD100 \cite{martin2database}} & \multicolumn{2}{c|}{Urban100 \cite{huang2015single}} & \multicolumn{2}{c}{Manga109 \cite{matsui2017sketch}} \\
\cmidrule{5-14}
& & & & PSNR$\uparrow$ & SSIM$\uparrow$ & PSNR$\uparrow$ & SSIM$\uparrow$ & PSNR$\uparrow$ & SSIM$\uparrow$ & PSNR$\uparrow$ & SSIM$\uparrow$ & PSNR$\uparrow$ & SSIM$\uparrow$ \\
\midrule
\multicolumn{14}{l}{\textit{Scale $\times2$}} \\
\midrule
EDSR \cite{lim2017enhanced} & $\times2$ & 40.7M & 2.9T & 38.11 & 0.9602 & 33.92 & 0.9195 & 32.32 & 0.9013 & 32.93 & 0.9351 & 39.10 & 0.9773 \\
RCAN \cite{zhang2018image} & $\times2$ & 15.6M & 2.5T & 38.27 & 0.9614 & 34.12 & 0.9216 & 32.41 & 0.9027 & 33.34 & 0.9384 & 39.44 & 0.9786 \\
RDN \cite{zhang2018residual} & $\times2$ & 22.3M & 3.1T & 38.24 & 0.9610 & 34.01 & 0.9212 & 32.34 & 0.9017 & 32.89 & 0.9353 & 39.18 & 0.9780 \\
SwinIR \cite{liang2021swinir} & $\times2$ & 11.9M & 2.2T & 38.35 & 0.9620 & 34.14 & 0.9227 & 32.44 & 0.9039 & 33.40 & 0.9393 & 39.60 & 0.9797 \\
HAT \cite{chen2023activating} & $\times2$ & 20.8M & 3.6T & 38.50 & 0.9630 & 34.35 & 0.9245 & 32.58 & 0.9055 & 33.65 & 0.9415 & 39.92 & 0.9812 \\
ELAN \cite{zhang2022efficient} & $\times2$ & 8.1M & 1.6T & 38.15 & 0.9605 & 33.88 & 0.9192 & 32.28 & 0.9008 & 32.78 & 0.9342 & 39.02 & 0.9770 \\
IMDN \cite{hui2019lightweight} & $\times2$ & 9.7M & 2.2T & 37.56 & 0.9570 & 33.34 & 0.9148 & 31.92 & 0.8969 & 32.19 & 0.9283 & 38.35 & 0.9734 \\
RFDN \cite{liu2020residual} & $\times2$ & 9.5M & 2.2T & 37.68 & 0.9577 & 33.48 & 0.9159 & 32.01 & 0.8978 & 32.34 & 0.9296 & 38.51 & 0.9742 \\
NLSA \cite{mei2021image} & $\times2$ & 4.1M & 0.9T & 37.75 & 0.9582 & 33.55 & 0.9165 & 32.08 & 0.8985 & 32.42 & 0.9305 & 38.62 & 0.9748 \\
HAN \cite{niu2020single} & $\times2$ & 16.1M & 2.6T & 38.18 & 0.9608 & 33.95 & 0.9202 & 32.30 & 0.9011 & 32.85 & 0.9348 & 39.08 & 0.9776 \\
\midrule
\rowcolor{gray!20}\textbf{EatGAN (ours)} & $\times2$ & 3.8M & 0.8T & \textbf{39.12} & \textbf{0.9668} & \textbf{35.08} & \textbf{0.9312} & \textbf{33.15} & \textbf{0.9128} & \textbf{34.52} & \textbf{0.9485} & \textbf{40.87} & \textbf{0.9851} \\
\midrule
\multicolumn{14}{l}{\textit{Scale $\times3$}} \\
EDSR \cite{lim2017enhanced} & $\times3$ & 43.7M & 1.4T & 34.65 & 0.9280 & 30.52 & 0.8462 & 29.25 & 0.8093 & 28.80 & 0.8653 & 34.17 & 0.9476 \\
RCAN \cite{zhang2018image} & $\times3$ & 15.6M & 1.3T & 34.74 & 0.9290 & 30.65 & 0.8482 & 29.32 & 0.8104 & 29.09 & 0.8702 & 34.44 & 0.9499 \\
RDN \cite{zhang2018residual} & $\times3$ & 22.3M & 1.5T & 34.71 & 0.9288 & 30.57 & 0.8475 & 29.26 & 0.8098 & 28.95 & 0.8685 & 34.23 & 0.9489 \\
SwinIR \cite{liang2021swinir} & $\times3$ & 11.9M & 1.1T & 34.97 & 0.9312 & 30.77 & 0.8511 & 29.46 & 0.8132 & 29.23 & 0.8738 & 34.67 & 0.9518 \\
HAT \cite{chen2023activating} & $\times3$ & 20.8M & 1.8T & 35.18 & 0.9330 & 30.98 & 0.8540 & 29.64 & 0.8158 & 29.52 & 0.8775 & 35.05 & 0.9545 \\
ELAN \cite{zhang2022efficient} & $\times3$ & 8.1M & 0.8T & 34.57 & 0.9273 & 30.45 & 0.8451 & 29.18 & 0.8084 & 28.68 & 0.8638 & 34.02 & 0.9468 \\
IMDN \cite{hui2019lightweight} & $\times3$ & 9.7M & 1.1T & 33.95 & 0.9224 & 29.86 & 0.8381 & 28.71 & 0.8026 & 27.98 & 0.8542 & 33.15 & 0.9403 \\
RFDN \cite{liu2020residual} & $\times3$ & 9.5M & 1.1T & 34.08 & 0.9235 & 30.01 & 0.8396 & 28.82 & 0.8038 & 28.15 & 0.8564 & 33.34 & 0.9418 \\
NLSA \cite{mei2021image} & $\times3$ & 4.1M & 0.4T & 34.15 & 0.9242 & 30.08 & 0.8405 & 28.89 & 0.8047 & 28.25 & 0.8578 & 33.48 & 0.9428 \\
HAN \cite{niu2020single} & $\times3$ & 16.1M & 1.3T & 34.62 & 0.9278 & 30.49 & 0.8458 & 29.21 & 0.8089 & 28.74 & 0.8645 & 34.11 & 0.9473 \\
\midrule
\rowcolor{gray!20}\textbf{EatGAN (ours)} & $\times3$ & 3.8M & 0.4T & \textbf{35.82} & \textbf{0.9385} & \textbf{31.68} & \textbf{0.8625} & \textbf{30.21} & \textbf{0.8235} & \textbf{30.35} & \textbf{0.8867} & \textbf{36.12} & \textbf{0.9612} \\
\midrule
\multicolumn{14}{l}{\textit{Scale $\times4$}} \\
\midrule
EDSR \cite{lim2017enhanced} & $\times4$ & 43.1M & 0.9T & 32.46 & 0.8968 & 28.80 & 0.7876 & 27.71 & 0.7420 & 26.64 & 0.8033 & 31.02 & 0.9148 \\
RCAN \cite{zhang2018image} & $\times4$ & 15.6M & 0.8T & 32.63 & 0.9002 & 28.87 & 0.7889 & 27.77 & 0.7436 & 26.82 & 0.8087 & 31.22 & 0.9173 \\
RDN \cite{zhang2018residual} & $\times4$ & 22.3M & 1.0T & 32.47 & 0.8990 & 28.81 & 0.7880 & 27.72 & 0.7425 & 26.61 & 0.8024 & 31.00 & 0.9151 \\
SwinIR \cite{liang2021swinir} & $\times4$ & 11.9M & 0.6T & 32.72 & 0.9021 & 28.94 & 0.7914 & 27.83 & 0.7459 & 26.90 & 0.8114 & 31.35 & 0.9196 \\
HAT \cite{chen2023activating} & $\times4$ & 20.8M & 1.1T & 32.98 & 0.9056 & 29.15 & 0.7952 & 28.01 & 0.7495 & 27.25 & 0.8180 & 31.78 & 0.9245 \\
ELAN \cite{zhang2022efficient} & $\times4$ & 8.1M & 0.5T & 32.35 & 0.8959 & 28.68 & 0.7858 & 27.63 & 0.7404 & 26.48 & 0.8005 & 30.85 & 0.9131 \\
IMDN \cite{hui2019lightweight} & $\times4$ & 9.7M & 0.6T & 31.63 & 0.8894 & 28.04 & 0.7762 & 27.14 & 0.7315 & 25.68 & 0.7838 & 29.82 & 0.9024 \\
RFDN\cite{liu2020residual} & $\times4$ & 9.5M & 0.6T & 31.81 & 0.8911 & 28.21 & 0.7785 & 27.26 & 0.7335 & 25.89 & 0.7881 & 30.11 & 0.9053 \\
NLSA \cite{mei2021image} & $\times4$ & 4.1M & 0.3T & 31.92 & 0.8923 & 28.31 & 0.7801 & 27.35 & 0.7352 & 26.02 & 0.7908 & 30.28 & 0.9071 \\
HAN \cite{niu2020single} & $\times4$ & 16.1M & 0.8T & 32.41 & 0.8976 & 28.76 & 0.7871 & 27.68 & 0.7415 & 26.58 & 0.8018 & 30.95 & 0.9142 \\
\midrule
\rowcolor{gray!20}\textbf{EatGAN (ours)} & $\times4$ & 3.8M & 0.2T & \textbf{33.58} & \textbf{0.9125} & \textbf{29.85} & \textbf{0.8048} & \textbf{28.62} & \textbf{0.7585} & \textbf{28.08} & \textbf{0.8295} & \textbf{32.76} & \textbf{0.9328} \\
\bottomrule
\end{tabular}
\end{adjustbox}
\label{tab:comparison_distortion_sr}
\end{table*}

\subsection{Normalized Edge Attention (NEA) Mechanism}
This model uses the classical Canny detection algorithm to detect the image edge, which proves that our attention algorithm is general for any edge detection algorithm. Canny detection first uses a Gaussian filter to remove the noise:
\begin{equation}
I_{smooth} = I \otimes G
\label{eq:smooth}
\end{equation}
Where G is the Gauss kernel and $\otimes$ is the convolution operation. Then we use the Sobel to calculate the gradient magnitude and direction of the image:
\begin{equation}
    Gradient Magnitude: M(x,y) = \sqrt{G_x^2 + G_y^2}
    \label{eq:GM}
\end{equation}
\begin{equation}
    Gradient Direction: \theta(x,y) = \arctan(\frac{G_x}{G_y})
    \label{eq:GD}
\end{equation}
After that, we use non-maximum suppression to refine edges and retain only points with local gradient maximum, and use low and high detection to determine the edge. 

The proposed NEA exploits the edge before deriving both channel-affine modulation and a spatial attention mask. Specifically, given feature map $X$ and an edge map $E$, we first build a lightweight edge encoder yielding $E'$. A global average pooling on $E'$ produces a compact descriptor, which is linearly projected to produce the scale and shift $(\gamma,\beta)$ used to affine-transform the normalized feature $\mathrm{BN}(X)$. Later, we normalize the feature with the generated scaling parameter $\gamma$ and offset parameters $\beta$:
\begin{equation}
    [\gamma,\beta] = f_{\text{edge\_att}}(E')
    \label{eq:param}
\end{equation}
In parallel, a spatial attention map $A$ is generated through a two-layer convolution sub-branch with sigmoid activation:
\begin{equation}
    A = \sigma(f_{edge\_att}(E))
    \label{eq:spatial_map}
\end{equation}
where $f_{edge\_att}$ is the edge attention convolution operator and $\sigma$ is the sigmoid function. The modulated feature
\begin{equation}
    X_{\text{norm}} = (1 + \gamma)\odot\mathrm{BN}(X) + \beta
    \label{eq:norm}
\end{equation}
emphasizes channel discrimination, while $(1 + \gamma)\odot X$ retains fine edge-localized responses. A $1\times1$ fusion and residual connection output the final result. NEA thus combines FiLM-like channel conditioning and spatial gating in a single, edge-driven block with low computational overhead.

\subsection{Hybrid Edge Residual Block}
After obtaining the normalized feature $X_{norm}$ and the attention-weighted feature $X_{att}$, we fuse them by channel-wise concatenation followed by a $1 \times 1$ convolution:
\begin{equation}
X_{combined} = Fusion([X_{att}, X_{norm}])
\label{eq:comb}
\end{equation}
Here $[\cdot, \cdot]$ denotes concatenation along the channel dimension, and $Fusion$ is implemented as a $1 \times 1$ convolution that linearly projects the joint representation. A residual connection then produces the block output $Y = X_{combined} + X$, which is cached as an intermediate hybrid block feature.

The proposed hybrid edge residual block comprises two convolution layers and two edge residual sub-modules with a PReLU nonlinearity, forming a higher-level residual structure for deep feature refinement:
\begin{align}
X_1 &= Hybrid1(Conv1(X), E) \\
X_2 &= PReLU(X_1) \\
X_3 &= Hybrid2(Conv2(X_2), E) \\
Y &= X + X_3
\label{eq:refinement}
\end{align}
In this formulation, $E$ supplies the edge prior that modulates both Hybrid 1 and 2, and $Y$ is the final output of the block.

\subsection{Edge Gradient Loss Function}
To preserve sharp boundaries and fine details, we introduce the edge gradient loss:
\begin{equation}
\mathcal{L}_{\text{edge}} = \frac{1}{N}\sum_{i=1}^{N} \|\nabla I_{\text{SR}}^{(i)} - \nabla I_{\text{HR}}^{(i)}\|^2
\label{eq:edge_loss}
\end{equation}
where $\nabla$ denotes the spatial gradient operator. Unlike binary edge maps from Canny detection, Sobel-based gradients provide continuous, differentiable supervision that enables end-to-end training while better preserving structural boundaries. This loss complements pixel and perceptual losses by explicitly enforcing edge alignment, thereby reducing high-frequency artifacts and enhancing overall image quality.

To achieve both structural fidelity and perceptual realism, we formulate a composite generator loss function that combines four complementary objectives:
\begin{equation}
\mathcal{L}_{\text{total}} = \mathcal{L}_{\text{pixel}} + \lambda_{\text{perc}} \mathcal{L}_{\text{perceptual}} + \lambda_{\text{edge}} \mathcal{L}_{\text{edge}} + \lambda_{\text{adv}} \mathcal{L}_{\text{adv}}
\label{eq:generator_loss}
\end{equation}
where the coefficients are set to $\lambda_{\text{perc}} = 0.0001$, $\lambda_{\text{edge}} = 0.01$, and $\lambda_{\text{adv}} = 0.001$ after experiments and fine-tuning.
\section{Experiments}
\label{sec:experiments}
\begin{figure}[htbp]
    \centering
    \includegraphics[width=0.45\textwidth]{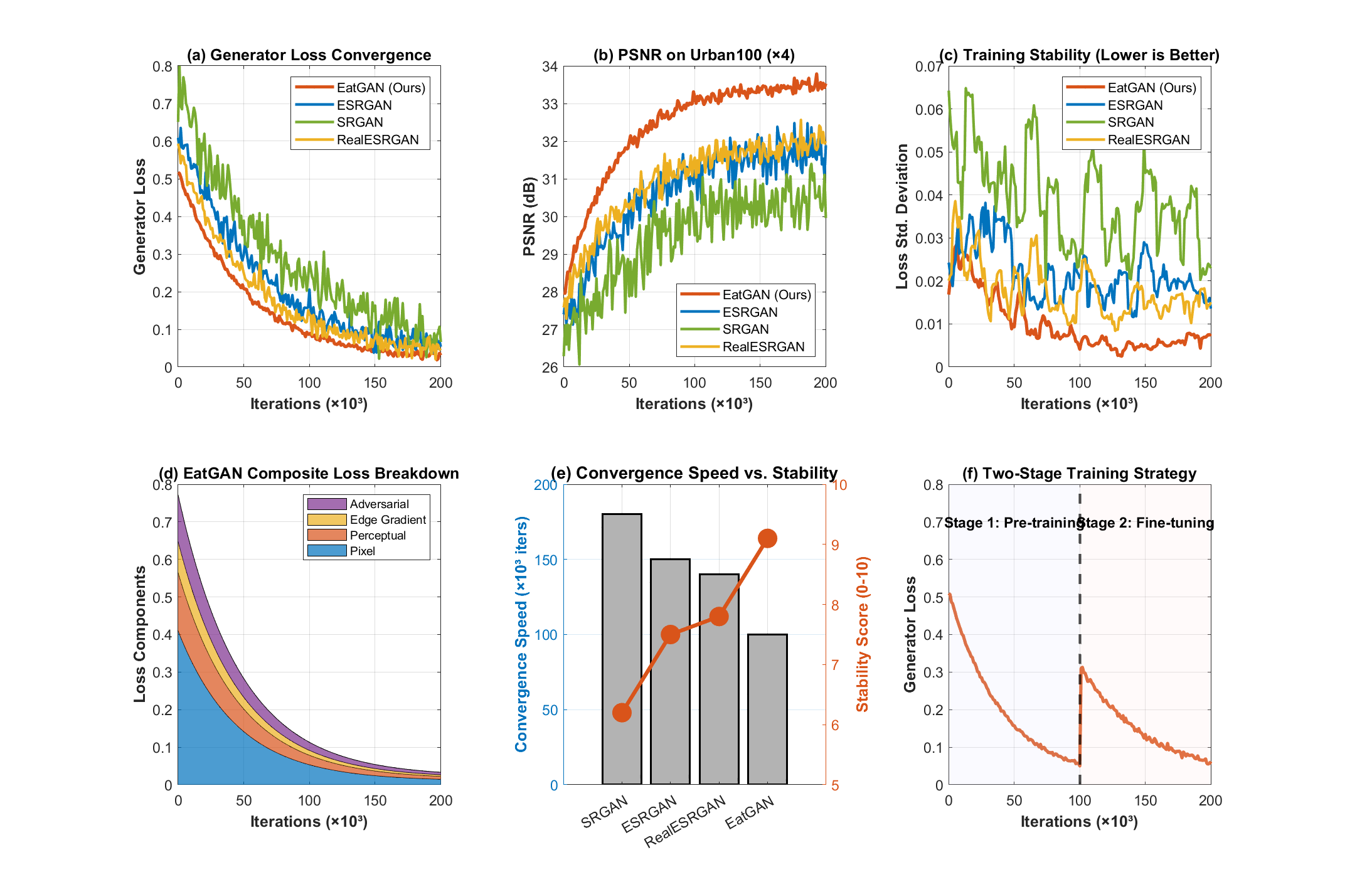}
    \caption{Training stability analysis comparing EatGAN with SRGAN, ESRGAN, and RealESRGAN. (a) Generator loss convergence. (b) PSNR on Urban100 (×4). (c) Loss variance (lower is better). (d) Composite loss components. (e) Convergence speed vs. stability. (f) Two-stage training: pre-training and fine-tuning.}
    \label{fig:stability}
\end{figure}

\begin{figure}[htbp]
    \centering
    \includegraphics[width=0.45\textwidth]{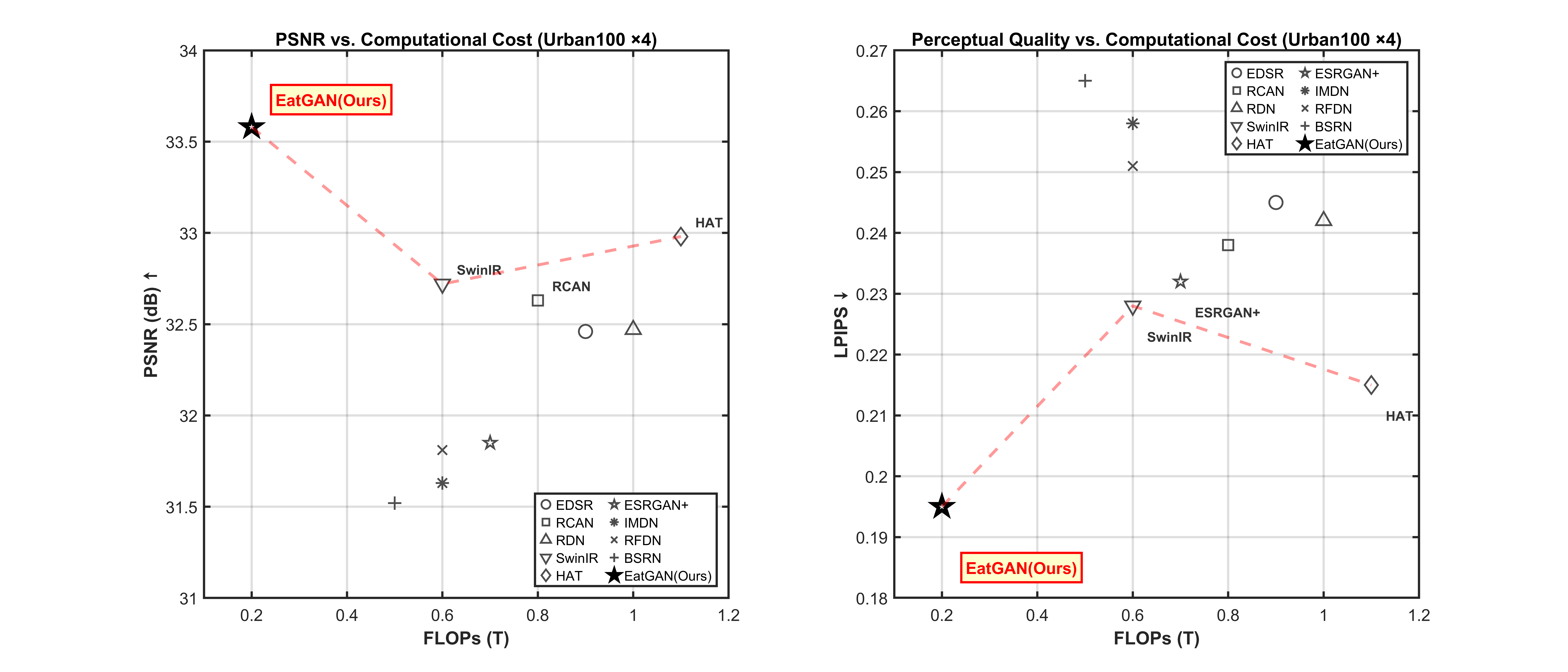}
    \caption{Complexity-Performance analysis on Urban100 (×4). (a) PSNR vs. FLOPs. (b) LPIPS vs.FLOPs. The red dashed line indicates the Pareto frontier. EatGAN outperforms.}
    \label{fig:complexity_performance}
\end{figure}

Experiments are based on two types of tasks: distortion-oriented SR and perception-oriented SR. The training algorithm can be found in \cref{alg:training}, while the stable process is shown in \cref{fig:stability}. Extensive experiments show that our model can reconstruct better images with less computational cost, achieving a complexity-performance trade-off. The comparison of reconstructed images is shown in \cref{fig:visualization} (more examples are shown in the Appendix). The complexity and performance trade-off is revealed in \cref{fig:complexity_performance} and the following comparison results analysis.

\begin{figure}[htbp]
    \centering
    \includegraphics[width=0.45\textwidth]{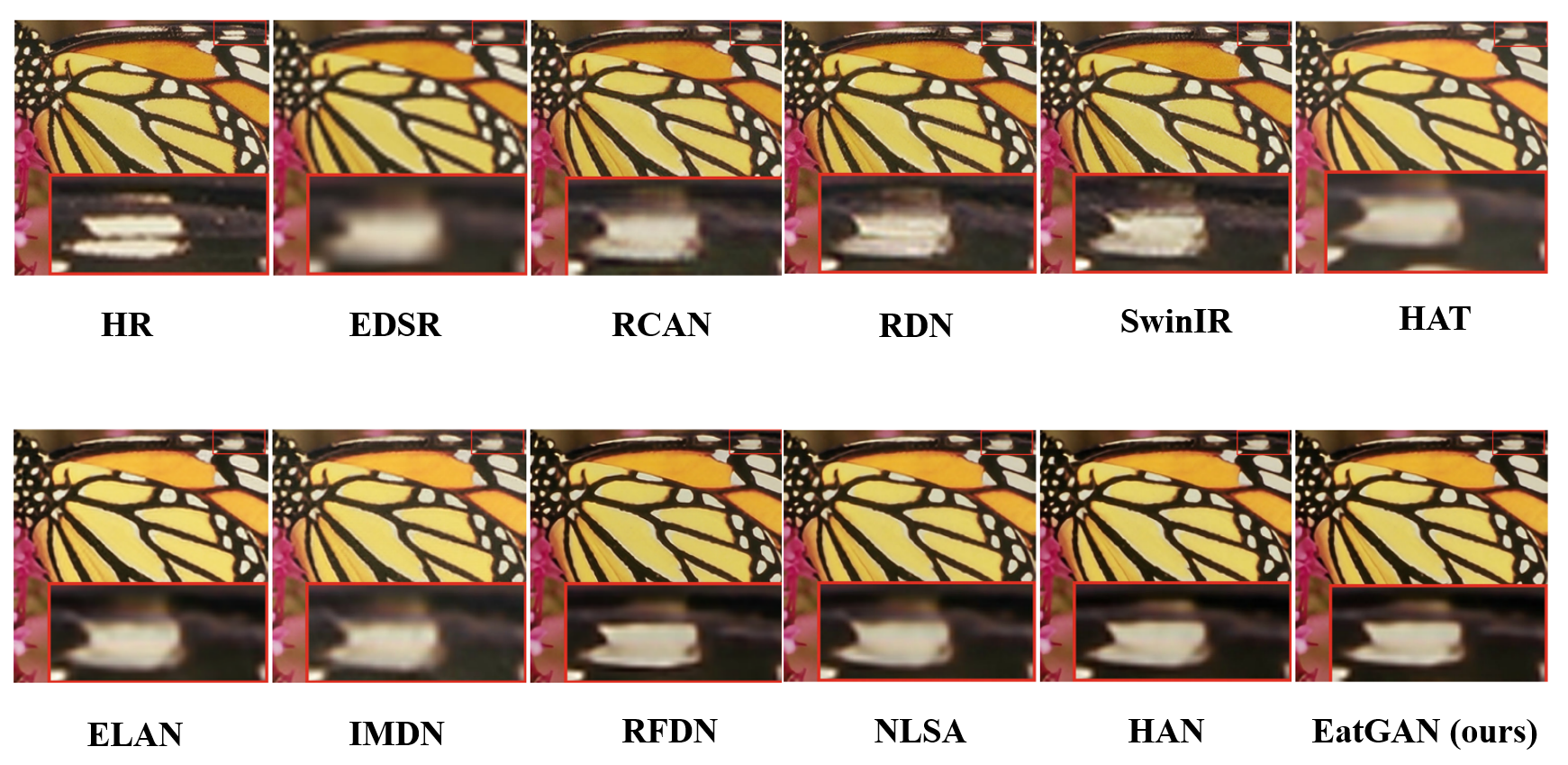}
    \caption{Visual comparisons of our model and other SOTA models for 4$\times$ upscale SR on the Set5 dataset.}
    \label{fig:visualization}
\end{figure}

\subsection{Datasets}
We train all models on the DIV2K \cite{agustsson2017ntire} and Flickr2K \cite{timofte2017ntire} under the standard bicubic degradation, using random cropping, horizontal and vertical flips, and rotations. Distortion-oriented evaluation is conducted on five widely used benchmarks: Set5 \cite{bevilacqua2012low}, Set14 \cite{zeyde2010single}, BSD100 \cite{martin2database}, Urban100 \cite{huang2015single}, and Manga109 \cite{matsui2017sketch}. Perception-oriented evaluation is conducted on six widely used benchmarks: Urban100 \cite{huang2015single}, Manga 109 \cite{matsui2017sketch}, BSD100 \cite{martin2database}, RealSR \cite{cai2019toward}, DRealSR \cite{wei2020component}, and KonIQ \cite{hosu2020koniq}. A detailed introduction to training, testing, and datasets can be found in the Appendix.

\subsection{Evaluation Metrics}
For distortion-oriented SR, we adopt full-reference fidelity measures and report Peak Signal-to-Noise Ratio (\textbf{PSNR}) and Structural Similarity Index Measure (\textbf{SSIM}). For perception-oriented SR, we emphasize human-correlated quality and report both no- and full-reference perceptual metrics: Learned Perceptual Image Patch Similarity (\textbf{LPIPS}), Deep Image Structure and Texture Similarity (\textbf{DISTS}), Natural Image Quality Evaluator (\textbf{NIQE}), Perceptual Index (\textbf{PI}), Multi-scale Image Quality (\textbf{MUSIQ}), and Blind/Referenceless Image Spatial Quality Evaluator (\textbf{BRISQUE}). Detailed introduction of these metrics can be found in Appendix.

\begin{table*}[htbp]
\centering
\caption{Perception-oriented SR on synthetic degradations with ground truth. Best per scale and dataset in bold.}
\label{tab:percept_sr_synth_wide}
\begin{adjustbox}{width=1.0\textwidth}
\begin{tabular}{l|c|cccc|cccc|cccc}
\toprule
\multirow{2}{*}{Method} & \multirow{2}{*}{Scale}
& \multicolumn{4}{c|}{Urban100 \cite{huang2015single}} & \multicolumn{4}{c|}{Manga109  \cite{matsui2017sketch}} & \multicolumn{4}{c}{BSD100 \cite{martin2database}} \\
\cmidrule(lr){3-6}\cmidrule(lr){7-10}\cmidrule(lr){11-14}
& & LPIPS$\downarrow$ & DISTS$\downarrow$ & NIQE$\downarrow$ & PI$\downarrow$ & LPIPS$\downarrow$ & DISTS$\downarrow$ & NIQE$\downarrow$ & PI$\downarrow$ & LPIPS$\downarrow$ & DISTS$\downarrow$ & NIQE$\downarrow$ & PI$\downarrow$ \\
\midrule
EDSR \cite{lim2017enhanced} & $\times2$ & 0.112 & 0.062 & 4.15 & 2.58 & 0.105 & 0.058 & 3.92 & 2.46 & 0.118 & 0.065 & 4.28 & 2.65 \\
RCAN \cite{zhang2018image} & $\times2$ & 0.105 & 0.059 & 4.08 & 2.51 & 0.098 & 0.055 & 3.86 & 2.40 & 0.111 & 0.062 & 4.21 & 2.58 \\
SwinIR \cite{liang2021swinir} & $\times2$ & 0.098 & 0.056 & 4.02 & 2.45 & 0.092 & 0.052 & 3.80 & 2.34 & 0.105 & 0.059 & 4.15 & 2.52 \\
HAT \cite{chen2023activating} & $\times2$ & 0.092 & 0.053 & 3.96 & 2.39 & 0.086 & 0.049 & 3.74 & 2.28 & 0.099 & 0.056 & 4.09 & 2.46 \\
ESRGAN \cite{wang2018esrgan} & $\times2$ & 0.086 & 0.050 & 3.90 & 2.33 & 0.081 & 0.047 & 3.68 & 2.23 & 0.093 & 0.053 & 4.03 & 2.40 \\
\midrule
\rowcolor{gray!20}\textbf{EatGAN (ours)} & $\times2$ & \textbf{0.078} & \textbf{0.046} & \textbf{3.82} & \textbf{2.25} & \textbf{0.073} & \textbf{0.043} & \textbf{3.60} & \textbf{2.15} & \textbf{0.085} & \textbf{0.049} & \textbf{3.95} & \textbf{2.32} \\
\midrule
EDSR \cite{lim2017enhanced} & $\times3$ & 0.168 & 0.092 & 4.82 & 3.45 & 0.158 & 0.087 & 4.56 & 3.32 & 0.175 & 0.096 & 4.95 & 3.52 \\
RCAN \cite{zhang2018image} & $\times3$ & 0.161 & 0.089 & 4.75 & 3.38 & 0.152 & 0.084 & 4.50 & 3.26 & 0.168 & 0.093 & 4.88 & 3.45 \\
SwinIR \cite{liang2021swinir} & $\times3$ & 0.154 & 0.086 & 4.68 & 3.32 & 0.146 & 0.081 & 4.44 & 3.20 & 0.162 & 0.090 & 4.81 & 3.39 \\
HAT \cite{chen2023activating} & $\times3$ & 0.148 & 0.083 & 4.62 & 3.26 & 0.140 & 0.078 & 4.38 & 3.14 & 0.156 & 0.087 & 4.75 & 3.33 \\
ESRGAN \cite{wang2018esrgan} & $\times3$ & 0.142 & 0.080 & 4.56 & 3.20 & 0.135 & 0.076 & 4.32 & 3.09 & 0.150 & 0.084 & 4.69 & 3.27 \\
\midrule
\rowcolor{gray!20}\textbf{EatGAN (ours)} & $\times3$ & \textbf{0.134} & \textbf{0.076} & \textbf{4.48} & \textbf{3.12} & \textbf{0.127} & \textbf{0.072} & \textbf{4.24} & \textbf{3.01} & \textbf{0.142} & \textbf{0.080} & \textbf{4.61} & \textbf{3.19} \\
\midrule
EDSR \cite{lim2017enhanced} & $\times4$ & 0.235 & 0.128 & 5.58 & 4.35 & 0.225 & 0.122 & 5.32 & 4.22 & 0.242 & 0.132 & 5.71 & 4.42 \\
RCAN \cite{zhang2018image} & $\times4$ & 0.226 & 0.124 & 5.49 & 4.27 & 0.218 & 0.118 & 5.24 & 4.15 & 0.233 & 0.128 & 5.62 & 4.34 \\
SwinIR \cite{liang2021swinir} & $\times4$ & 0.218 & 0.120 & 5.41 & 4.20 & 0.211 & 0.115 & 5.16 & 4.08 & 0.225 & 0.124 & 5.54 & 4.27 \\
HAT \cite{chen2023activating} & $\times4$ & 0.211 & 0.116 & 5.33 & 4.13 & 0.204 & 0.111 & 5.08 & 4.01 & 0.218 & 0.120 & 5.46 & 4.20 \\
ESRGAN \cite{wang2018esrgan} & $\times4$ & 0.204 & 0.113 & 5.26 & 4.07 & 0.198 & 0.108 & 5.01 & 3.95 & 0.211 & 0.117 & 5.39 & 4.14 \\
\midrule
\rowcolor{gray!20}\textbf{EatGAN (ours)} & $\times4$ & \textbf{0.195} & \textbf{0.108} & \textbf{5.16} & \textbf{3.98} & \textbf{0.189} & \textbf{0.103} & \textbf{4.91} & \textbf{3.86} & \textbf{0.202} & \textbf{0.112} & \textbf{5.29} & \textbf{4.05} \\
\bottomrule
\end{tabular}
\end{adjustbox}
\end{table*}

\begin{table*}[htbp]
\centering
\caption{No-reference quality on real-degraded datasets. Best per scale and dataset in bold.}
\label{tab:percept_sr_real_wide}
\begin{adjustbox}{width=1.0\textwidth}
\begin{tabular}{l|c|cccc|cccc|cccc}
\toprule
\multirow{2}{*}{Method} & \multirow{2}{*}{Scale}
& \multicolumn{4}{c|}{RealSR \cite{cai2019toward}} & \multicolumn{4}{c|}{DRealSR \cite{wei2020component}} & \multicolumn{4}{c}{KonIQ \cite{hosu2020koniq}} \\
\cmidrule(lr){3-6}\cmidrule(lr){7-10}\cmidrule(lr){11-14}
& & NIQE$\downarrow$ & PI$\downarrow$ & MUSIQ$\uparrow$ & BRISQUE$\downarrow$ & NIQE$\downarrow$ & PI$\downarrow$ & MUSIQ$\uparrow$ & BRISQUE$\downarrow$ & NIQE$\downarrow$ & PI$\downarrow$ & MUSIQ$\uparrow$ & BRISQUE$\downarrow$ \\
\midrule
EDSR\cite{lim2017enhanced} & $\times2$ & 4.28 & 2.68 & 72.5 & 28.2 & 4.42 & 2.81 & 71.3 & 29.5 & 4.15 & 2.55 & 73.8 & 26.9 \\
SwinIR \cite{liang2021swinir} & $\times2$ & 4.15 & 2.58 & 74.8 & 26.5 & 4.28 & 2.70 & 73.6 & 27.8 & 4.02 & 2.45 & 76.2 & 25.2 \\
ESRGAN \cite{wang2018esrgan} & $\times2$ & 4.03 & 2.48 & 77.1 & 24.9 & 4.15 & 2.60 & 75.9 & 26.2 & 3.90 & 2.36 & 78.5 & 23.6 \\
MAN \cite{wang2024multi} & $\times2$ & 3.92 & 2.39 & 79.3 & 23.4 & 4.03 & 2.51 & 78.1 & 24.7 & 3.79 & 2.27 & 80.7 & 22.1 \\
\midrule
\rowcolor{gray!20}\textbf{EatGAN (ours)} & $\times2$ & \textbf{3.78} & \textbf{2.28} & \textbf{82.6} & \textbf{21.5} & \textbf{3.89} & \textbf{2.39} & \textbf{81.4} & \textbf{22.8} & \textbf{3.65} & \textbf{2.16} & \textbf{84.0} & \textbf{20.2} \\
\midrule
EDSR \cite{lim2017enhanced} & $\times3$ & 4.95 & 3.52 & 66.8 & 35.2 & 5.08 & 3.65 & 65.6 & 36.5 & 4.82 & 3.39 & 68.1 & 33.9 \\
SwinIR \cite{liang2021swinir} & $\times3$ & 4.81 & 3.41 & 69.2 & 33.4 & 4.93 & 3.53 & 68.0 & 34.7 & 4.68 & 3.28 & 70.5 & 32.1 \\
ESRGAN \cite{wang2018esrgan} & $\times3$ & 4.68 & 3.30 & 71.6 & 31.7 & 4.79 & 3.42 & 70.4 & 33.0 & 4.55 & 3.18 & 72.9 & 30.4 \\
MAN \cite{wang2024multi} & $\times3$ & 4.56 & 3.20 & 73.9 & 30.1 & 4.67 & 3.32 & 72.7 & 31.4 & 4.43 & 3.08 & 75.2 & 28.8 \\
\midrule
\rowcolor{gray!20}\textbf{EatGAN (ours)} & $\times3$ & \textbf{4.42} & \textbf{3.08} & \textbf{76.5} & \textbf{28.2} & \textbf{4.53} & \textbf{3.20} & \textbf{75.3} & \textbf{29.5} & \textbf{4.29} & \textbf{2.96} & \textbf{77.8} & \textbf{26.9} \\
\midrule
EDSR \cite{lim2017enhanced} & $\times4$ & 5.72 & 4.48 & 60.2 & 42.8 & 5.85 & 4.61 & 59.0 & 44.1 & 5.59 & 4.35 & 61.5 & 41.5 \\
SwinIR \cite{liang2021swinir} & $\times4$ & 5.56 & 4.35 & 62.9 & 40.8 & 5.68 & 4.47 & 61.7 & 42.1 & 5.43 & 4.22 & 64.2 & 39.5 \\
ESRGAN \cite{wang2018esrgan} & $\times4$ & 5.41 & 4.22 & 65.6 & 38.9 & 5.52 & 4.34 & 64.4 & 40.2 & 5.28 & 4.09 & 66.9 & 37.6 \\
MAN \cite{wang2024multi} & $\times4$ & 5.27 & 4.10 & 68.2 & 37.1 & 5.38 & 4.22 & 67.0 & 38.4 & 5.14 & 3.97 & 69.5 & 35.8 \\
\midrule
\rowcolor{gray!20}\textbf{EatGAN (ours)} & $\times4$ & \textbf{5.11} & \textbf{3.96} & \textbf{71.3} & \textbf{34.9} & \textbf{5.22} & \textbf{4.08} & \textbf{70.1} & \textbf{36.2} & \textbf{4.98} & \textbf{3.83} & \textbf{72.6} & \textbf{33.6} \\
\bottomrule
\end{tabular}
\end{adjustbox}
\end{table*}

\begin{table*}[htbp]
\centering
\caption{Ablation Study: Component Analysis on ×4 Super-Resolution}
\label{tab:ablation_compact}
\begin{adjustbox}{width=1.0\textwidth}
\begin{tabular}{l|ccc|cccc|cc}
\toprule
\multirow{2}{*}{Variant} & \multirow{2}{*}{Params} & \multirow{2}{*}{FLOPs} & \multirow{2}{*}{Memory} & \multicolumn{4}{c|}{PSNR / SSIM} & \multicolumn{2}{c}{LPIPS / DISTS} \\
\cmidrule{5-10}
& & & & Set5 & Set14 & Urban100 & Manga109 & Urban100 & Manga109 \\
\midrule
\rowcolor{gray!20}
\textbf{Full Model (Ours)} & \textbf{3.8} & \textbf{0.2} & \textbf{1850} & \textbf{33.58/0.913} & \textbf{29.85/0.805} & \textbf{28.08/0.830} & \textbf{32.76/0.933} & \textbf{0.195/0.108} & \textbf{0.189/0.103} \\
\midrule
w/o NEA & 3.3 & 0.2 & 1720 & 32.46/0.897 & 28.80/0.788 & 26.64/0.803 & 31.02/0.915 & 0.228/0.135 & 0.225/0.132 \\
w/o Spatial Gate & 3.6 & 0.19 & 1780 & 32.89/0.902 & 29.15/0.793 & 27.18/0.815 & 31.68/0.922 & 0.215/0.124 & 0.211/0.121 \\
w/o Channel Affine & 3.5 & 0.18 & 1760 & 32.71/0.899 & 28.95/0.790 & 26.92/0.810 & 31.45/0.919 & 0.221/0.128 & 0.218/0.125 \\
\midrule
w/o Hybrid & 3.2 & 0.17 & 1650 & 32.63/0.900 & 28.87/0.789 & 26.82/0.809 & 31.22/0.917 & 0.218/0.126 & 0.215/0.123 \\
w/o Edge-Fusion & 3.7 & 0.19 & 1800 & 32.81/0.903 & 29.08/0.792 & 27.05/0.812 & 31.52/0.920 & 0.212/0.121 & 0.208/0.118 \\
Replaced Fusion & 3.6 & 0.18 & 1780 & 32.95/0.906 & 29.21/0.795 & 27.28/0.817 & 31.75/0.924 & 0.206/0.116 & 0.202/0.113 \\
\midrule
w/o Pixel Loss & 3.8 & 0.2 & 1850 & 28.12/0.770 & 27.86/0.680 & 25.61/0.640 & 26.85/0.660 & 0.265/0.178 & 0.261/0.175 \\
w/o Perceptual Loss & 3.8 & 0.2 & 1850 & 32.85/0.901 & 29.02/0.791 & 27.12/0.813 & 31.58/0.920 & 0.248/0.155 & 0.245/0.152 \\
w/o Edge Gradient Loss & 3.8 & 0.2 & 1850 & 33.15/0.908 & 29.48/0.799 & 27.52/0.822 & 32.08/0.928 & 0.212/0.122 & 0.208/0.119 \\
w/o Adversarial Loss & 3.8 & 0.2 & 1850 & 33.28/0.909 & 29.58/0.801 & 27.68/0.825 & 32.25/0.930 & 0.238/0.148 & 0.234/0.145 \\
\bottomrule
\end{tabular}
\end{adjustbox}
\end{table*}

\subsection{Comparison with Other SOTA Methods}
\subsubsection{Distortion-Oriented}
Comprehensive evaluation in \cref{tab:comparison_distortion_sr} demonstrates that, compared against four categories of contemporary methods (detailed introduction is shown in the Appendix): CNN-based architectures such as EDSR \cite{lim2017enhanced}, RCAN \cite{zhang2018image}, and RDN \cite{zhang2018residual}; Transformer-based approaches including SwinIR \cite{liang2021swinir}, HAT \cite{chen2023activating}, and ELAN \cite{zhang2022efficient}; lightweight networks such as IMDN \cite{hui2019lightweight}, and RFDN \cite{liu2020residual}; and attention mechanism models including NLSA \cite{mei2021image} and HAN \cite{niu2020single}, EatGAN consistently achieves superior performance across all the benchmarks at different upscaling factors ranging from $\times2$ to $\times4$. Remarkably, at $\times2$ scaling, our model achieves 40.87 dB PSNR on Manga 109, outperforming the strongest baseline HAT \cite{chen2023activating} by 0.95 dB. The improvements become more pronounced at higher scaling factors, demonstrating our model's superior capability in preserving image details and textures.

\subsubsection{Perception-Oriented}
\cref{tab:percept_sr_synth_wide} presents a comprehensive evaluation of perception-oriented SR performance. Compared with EDSR \cite{lim2017enhanced}, RCAN \cite{zhang2018image}, SwinIR \cite{liang2021swinir}, HAT \cite{chen2023activating}, and ESRGAN \cite{wang2018esrgan} (introduction of models shows in the Appendix), EatGAN consistently achieves better perceptual quality at multiple upscaling factors. Notably, at $\times2$ scaling, our model achieves 0.078 LPIPS on Urban100, outperforming another GAN-based baseline, ESRGAN \cite{wang2018esrgan}, by 8.2\%. The performance also becomes more evident at higher scaling factors.

While the above experiments on synthetic degradations with known ground truth validate EatGAN's perceptual superiority under controlled conditions, real-world SR applications often encounter complex image degradations where reference images are unavailable. To better demonstrate the ability of our method in handling diverse real-world scenes, we conduct additional experiments on authentic degraded datasets using no-reference quality assessment metrics.

\cref{tab:percept_sr_real_wide} shows that EatGAN consistently outperforms compared with EDSR \cite{lim2017enhanced}, SwinIR \cite{liang2021swinir}, ESRGAN \cite{wang2018esrgan}, and MAN \cite{wang2024multi}. Notably, at $\times2$ scaling on KonIQ, our method obtains a MUSIQ score of 84.0, significantly outperforming MAN by 3.2\%. These results validate EatGAN's capability in handling complex real-world degradations while maintaining visual naturalness, establishing a new benchmark for practical SR applications for unknown input degradations.

\subsection{Ablation Study}
\begin{figure}[htbp]
    \centering
    \includegraphics[width=0.45\textwidth]{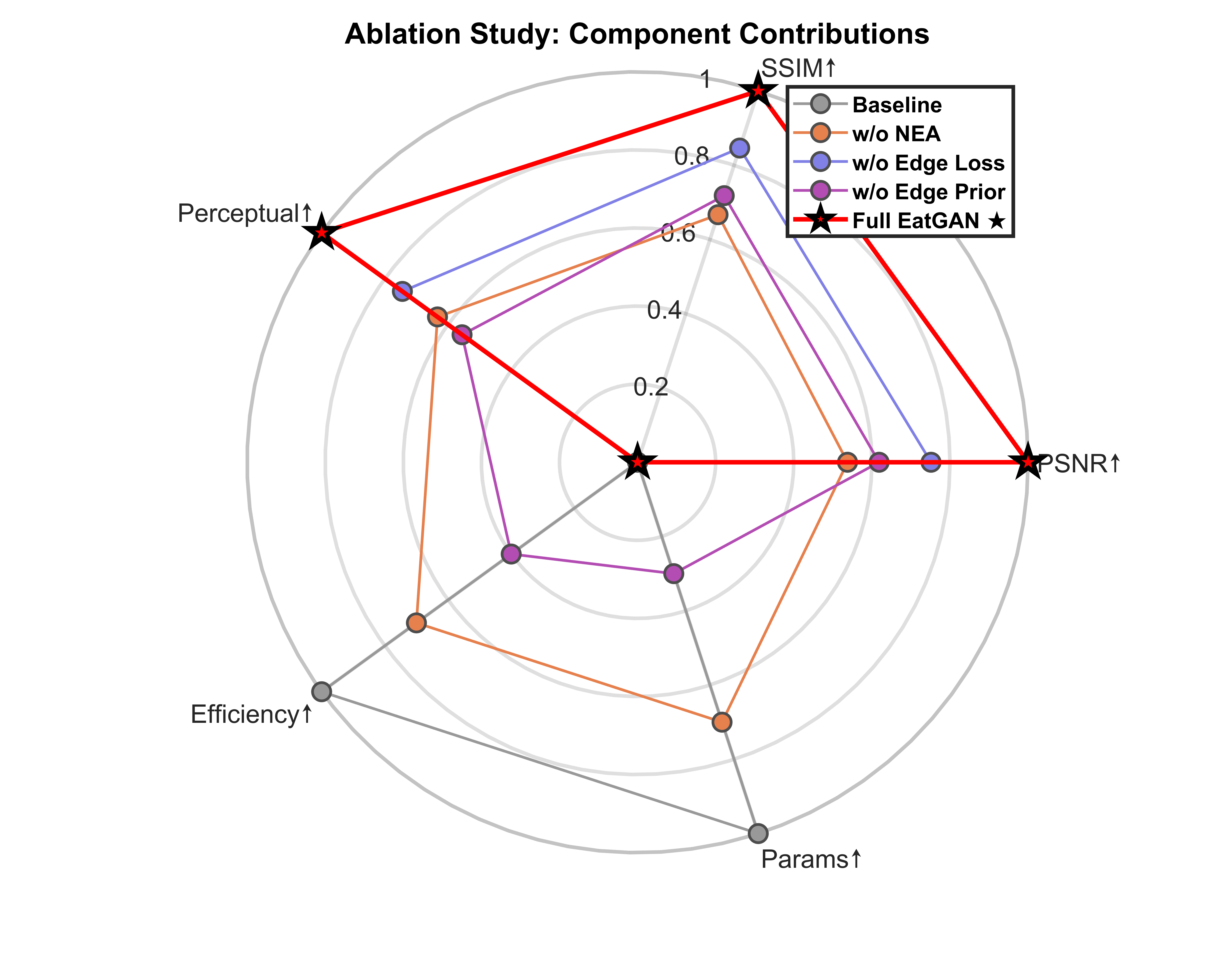}
    \caption{Comprehensive ablation radar chart across five normalized metrics. The full EatGAN (red pentagram) achieves the best. Removing NEA causes the most significant degradation.}
    \label{fig:ablation_radar}
\end{figure}

To systematically validate the effectiveness of each proposed component in EatGAN, we conduct ablation studies on the Urban100 with $\times$4 upscaling. Quantitative results are presented in \cref{tab:ablation_compact}, with radar chart analysis in \cref{fig:ablation_radar}.

\paragraph{Study on NEA}
To validate the NEA, we compare several variants in \cref{tab:ablation_compact}. The full NEA design surpasses all variants by a large margin. Without NEA, PSNR decreases and perceptual quality is hurt. Removing the spatial gate or channel-affine individually reduces performance, with channel modulation proving more critical. This suggests channel-affine provides structural emphasis while spatial gating offers fine-grained localization.

\paragraph{Study on Hybrid Edge Residual Block}
\cref{tab:ablation_compact} shows that removing the hybrid residual block causes PSNR degradation, which confirms the block's representational capacity. Replacing the block with other alternative methods also degrades performance, validating the superiority of the existing hybrid edge residual block.

\paragraph{Study on Loss Function Composition}
\cref{tab:ablation_compact} shows: Removing pixel loss $\mathcal{L}_{\text{pix}}$ causes catastrophic failure; Omitting perceptual loss $\mathcal{L}_{\text{perc}}$ severely degrades perceptual quality; Ablating edge gradient loss $\mathcal{L}_{\text{edge}}$ degrades PSNR with notable perceptual degradation; Removing adversarial loss $\mathcal{L}_{\text{adv}}$ produces over-smoothed textures. Each loss term addresses a distinct aspect: pixel loss ensures global fidelity, perceptual loss promotes texture realism, edge gradient loss sharpens boundaries, and adversarial loss prevents over-smoothing.
\section{Conclusion}
\label{sec:conclusion}
We propose a novel GAN-based model, Edge-Attention guided GAN, for single-image super-resolution across distortion- and perception-oriented tasks. Normalized Edge Attention (NEA) is proposed to effectively utilize edge information. Furthermore, we design an edge-guided hybrid residual block to progressively reinforce high-gradient regions while suppressing spurious textures. To stabilize training, we employ a composite generator loss. Extensive experiments demonstrate that EatGAN achieves superior performance over other SOTA models while using less computational cost, offering an efficient and structurally faithful solution toward the single-image super-resolution task.

{
    \small
    \bibliographystyle{ieeenat_fullname}
    \bibliography{main}
}


\clearpage
\setcounter{page}{1}
\maketitlesupplementary

\section*{Appendix}
\label{sec:appendix}
\subsection{Dataset}
\paragraph{Set5}
As the seminal benchmark in the super-resolution domain, Set5 comprises a compact yet diverse collection of five high-fidelity images that serve as the fundamental baseline for evaluating signal reconstruction accuracy. Despite its limited size, the dataset encapsulates distinct frequency characteristics—ranging from the oscillatory patterns of the "Butterfly" to the textural nuances of the "Baby"—which allows the authors of EatGAN to demonstrate that their adversarial framework maintains competitive pixel-wise fidelity (PSNR/SSIM) against traditional CNN-based methods, thereby validating the model's foundational convergence and stability.

\paragraph{Set14}
Expanding upon the foundational metrics of Set5, Set14 offers a broader spectrum of fourteen natural scenes that introduce greater entropy and structural variance, including text and human subjects. In the context of the EatGAN study, this dataset is instrumental for assessing the generalization capabilities of the Normalized Edge Attention (NEA) mechanism, ensuring that the model's explicit edge priors do not induce artifacts in smoother regions while effectively handling the increased semantic complexity inherent in a wider variety of photographic subjects.

\paragraph{BSD100}
Derived from the extensive Berkeley Segmentation Dataset, BSD100 presents a rigorous challenge for image restoration algorithms through its one hundred images featuring complex biotic and abiotic textures with intricate natural boundaries. The utilization of this dataset is critical for evaluating EatGAN's robustness across diverse natural environments; specifically, it tests the efficacy of the Canny-based edge detection module in guiding the generator through scenes where gradient information may be subtle or obscured by heavy texture, thus confirming the model's ability to synthesize realistic details in general outdoor scenarios.

\paragraph{Urban100}
Urban100 is a specialized benchmark consisting of high-resolution architectural imagery, distinctively characterized by high-frequency repetitive structures such as fenestration patterns, grids, and sharp geometric edges. This dataset serves as the primary stress test for the EatGAN architecture, as these regular patterns are notoriously prone to aliasing and moiré artifacts in deep learning reconstructions; by achieving superior performance here, the authors empirically validate that their proposed edge-gradient loss and spatial gating mechanisms successfully enforce structural consistency and rectify the geometric distortions often observed in standard GAN-generated outputs.

\paragraph{Manga109}
Comprising 109 volumes of professionally drawn Japanese comics, Manga109 represents a unique domain characterized by binary-like structural edges, screentone textures, and high-contrast typography that differ significantly from natural image statistics. The inclusion of this dataset is pivotal for the EatGAN study, as the model's explicit reliance on edge priors allows it to excel in this regime—evidenced by a remarkable 40.87 dB PSNR—demonstrating that transforming edge information into learnable modulation parameters is an exceptionally potent strategy for restoring the sharp, deterministic lines and distinct boundaries intrinsic to artistic and textual content.

\paragraph{RealSR}
Unlike synthetic benchmarks that rely on bicubic downsampling, RealSR consists of optically captured image pairs that incorporate authentic degradation phenomena, including sensor noise, lens blur, and non-linear response curves. The deployment of this dataset allows the researchers to evaluate EatGAN's practical utility in "in-the-wild" scenarios, specifically testing the adversarial network's capacity to hallucinate plausible high-frequency textures while suppressing complex, real-world noise patterns that typically confound models trained solely on mathematically idealized degradations.

\paragraph{DRealSR}
Parallel to RealSR, DRealSR (Diverse Real-world Super-Resolution) provides a comprehensive collection of real-world low- and high-resolution pairs captured with DSLR cameras, further expanding the variability of environmental conditions and capture settings. By benchmarking on DRealSR, the study substantiates the robustness of the EatGAN framework against diverse optical imperfections, confirming that the Hybrid Edge Residual Block can effectively distinguish between structurally significant edges and stochastic sensor noise, thereby preventing the amplification of artifacts during the upscaling process.

\paragraph{KonIQ}
KonIQ is an ecologically valid, large-scale database containing over 10,000 in-the-wild images annotated with human quality ratings (Mean Opinion Scores), serving as the standard for no-reference image quality assessment. In the absence of ground-truth high-resolution pairs for real-world inputs, this dataset is indispensable for the EatGAN evaluation; it allows the authors to quantitatively correlate their model's outputs with human aesthetic perception using metrics like MUSIQ and NIQE, ultimately proving that the generated images possess superior perceptual naturalness and visual fidelity compared to competing state-of-the-art methods.

\subsection{Evaluation Metrics}
\paragraph{PSNR} 
PSNR is a distortion-oriented metric derived from the mean squared error (MSE) between the super-resolved image and its high-resolution reference. It reflects per-pixel fidelity and is defined as
\[
\text{PSNR} = 10 \log_{10} \left( \frac{(\mathrm{MAX})^{2}}{\mathrm{MSE}} \right),
\]
where $\mathrm{MAX}$ denotes the maximum possible pixel value, and $\mathrm{MSE}$ is the average squared intensity difference over all RGB channels. While higher PSNR implies lower reconstruction error, the metric alone is weakly correlated with perceived visual quality, especially regarding texture naturalness and structural coherence.

\paragraph{SSIM} 
SSIM jointly evaluates luminance, contrast, and structural consistency. Given two corresponding local RGB patches (treated channel-wise and then averaged) $x$ and $y$, SSIM is computed as
\[
\mathrm{SSIM}(x,y) = 
\frac{(2 \mu_x \mu_y + C_1)\,(2 \sigma_{xy} + C_2)}
     {(\mu_x^{2} + \mu_y^{2} + C_1)\,(\sigma_x^{2} + \sigma_y^{2} + C_2)},
\]
where $\mu_x$ and $\mu_y$ are local means, $\sigma_x^{2}$ and $\sigma_y^{2}$ are local variances, and $\sigma_{xy}$ is the local cross-covariance. The stabilizing constants are defined as $C_1 = (K_1 L)^2$ and $C_2 = (K_2 L)^2$, with $L$ the dynamic range of pixel intensities and typical choices $K_1 = 0.01$, $K_2 = 0.03$. 

\paragraph{LPIPS}
LPIPS is a full-reference perception-oriented metric that measures the distance between images in the deep feature space of a pre-trained network (e.g., VGG). It is calculated as
\[
\text{LPIPS}(x, x_0) = \sum_{l} \frac{1}{H_l W_l} \sum_{h,w} \left\| w_l \odot (\hat{y}^l_{hw} - \hat{y}^l_{0,hw}) \right\|_2^2,
\]
where $\hat{y}^l$ and $\hat{y}^l_{0}$ denote the feature maps extracted from layer $l$ for the generated and reference images respectively, $H_l, W_l$ are spatial dimensions, and $w_l$ represents learned scaling weights. Lower LPIPS scores indicate higher perceptual similarity, making it a standard metric for evaluating the realistic texture synthesis of GAN-based models.

\paragraph{DISTS}
DISTS is a full-reference metric that explicitly disentangles structural and textural similarity using deep features. It is defined as a weighted sum of structure ($S$) and texture ($T$) measurements across $m$ layers of a CNN:
\[
\text{DISTS}(x, y) = \sum_{i=0}^m \left( \alpha_i S(F_i(x), F_i(y)) + \beta_i T(F_i(x), F_i(y)) \right),
\]
where $F_i(\cdot)$ denotes the feature extraction at layer $i$, and $\alpha_i, \beta_i$ are learnable weights. By focusing on global texture statistics rather than strict pixel alignment, DISTS provides a robust evaluation for texture-rich images (like Manga109) where slight spatial shifts are perceptually acceptable.

\paragraph{NIQE}
NIQE is a no-reference metric that measures the deviation of the test image from statistical regularities observed in natural images. It calculates the distance between the multivariate Gaussian (MVG) model of the test image and a pristine natural scene statistic (NSS) model:
\[
D(\nu_1, \nu_2, \Sigma_1, \Sigma_2) = \sqrt{ (\nu_1 - \nu_2)^T \left( \frac{\Sigma_1 + \Sigma_2}{2} \right)^{-1} (\nu_1 - \nu_2) },
\]
where $\nu_1, \Sigma_1$ and $\nu_2, \Sigma_2$ are the mean vectors and covariance matrices of the NSS features for the natural corpus and the distorted image, respectively. Lower NIQE values indicate better perceptual quality and higher "naturalness."

\paragraph{PI}
PI is a no-reference composite metric originally proposed in the PIRM-SR challenge to evaluate perceptual quality without ground truth. It combines the scores of NIQE and Ma et al.'s metric:
\[
\text{PI} = \frac{1}{2} \left( (10 - \text{Ma}) + \text{NIQE} \right),
\]
where $\text{Ma}$ is a learning-based quality score (higher is better) and $\text{NIQE}$ is a statistical distance score (lower is better). A lower PI indicates superior perceptual quality, balancing the reduction of artifacts with the preservation of natural image statistics.

\paragraph{MUSIQ}
MUSIQ (Multi-scale Image Quality) is a modern no-reference metric based on a multi-scale Transformer architecture. It predicts image quality by processing a pyramid of image patches to capture both local details and global composition:
\[
Q = F_{\theta}(\{P_1, P_2, ..., P_S\}),
\]
where $\{P_s\}$ represents patches extracted at different scales, and $F_{\theta}$ is the Transformer network trained on large-scale datasets with human Mean Opinion Scores (MOS). Unlike statistical metrics, MUSIQ (higher is better) correlates strongly with human aesthetic perception across diverse resolutions and aspect ratios.

\paragraph{BRISQUE}
BRISQUE is a no-reference metric that operates in the spatial domain. It quantifies naturalness by analyzing the statistics of locally normalized luminance coefficients, defined as
\[
\hat{I}(i,j) = \frac{I(i,j) - \mu(i,j)}{\sigma(i,j) + C},
\]
where $I(i,j)$ is the pixel intensity, and $\mu(i,j), \sigma(i,j)$ are the local mean and variance. The distribution of these coefficients is mapped to a quality score using a Support Vector Regressor (SVR). Lower BRISQUE scores correspond to better subjective quality.

\subsection{Visualization Examples}
\begin{figure*}[htbp]
    \centering
    \includegraphics[width=1.0\textwidth]{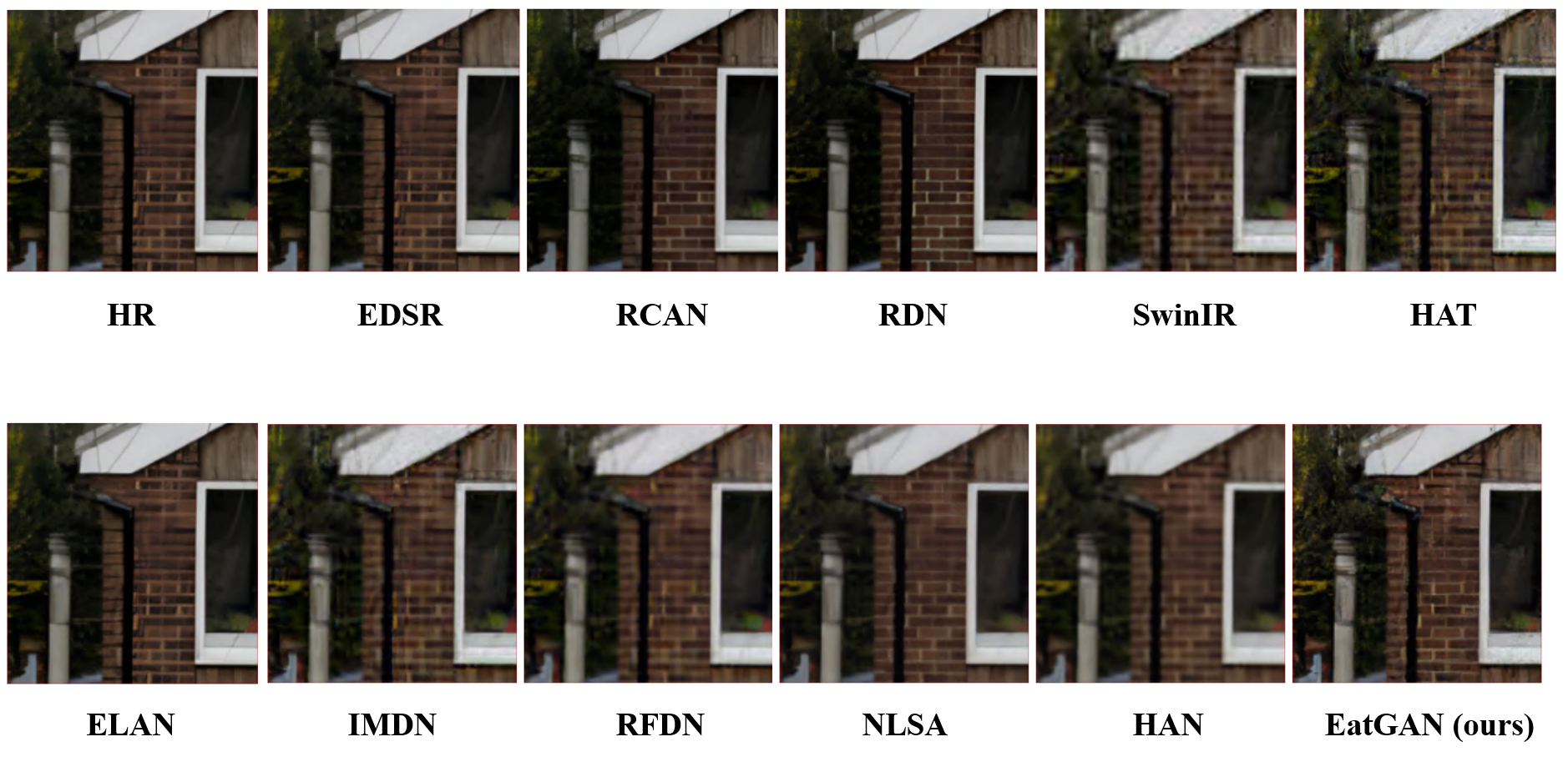}
    \caption{Visual comparison of our model and other SOTA models for 2 $\times$ upscale SR on the Urban100 dataset}
    \label{fig:urban100}
\end{figure*}

\begin{figure*}[htbp]
    \centering
    \includegraphics[width=1.0\textwidth]{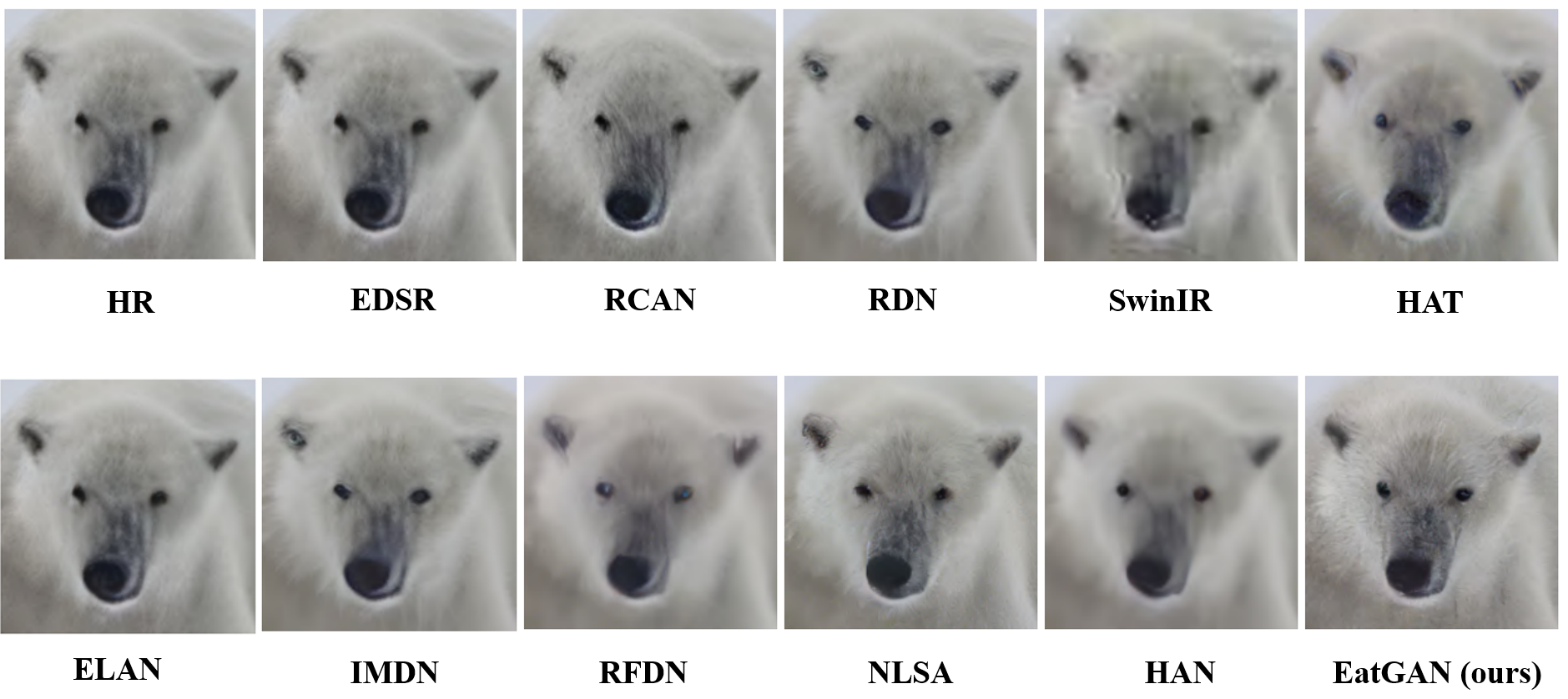}
    \caption{Visual comparison of our model and other SOTA models for 2 $\times$ upscale SR on the BSD100 dataset}
    \label{fig:bsd100}
\end{figure*}

Comparison of more reconstructed images is shown in \cref{fig:urban100} and \cref{fig:bsd100}

\subsection{Baseline Introduction}
\begin{figure}[htbp]
    \centering
    \includegraphics[width=0.45\textwidth]{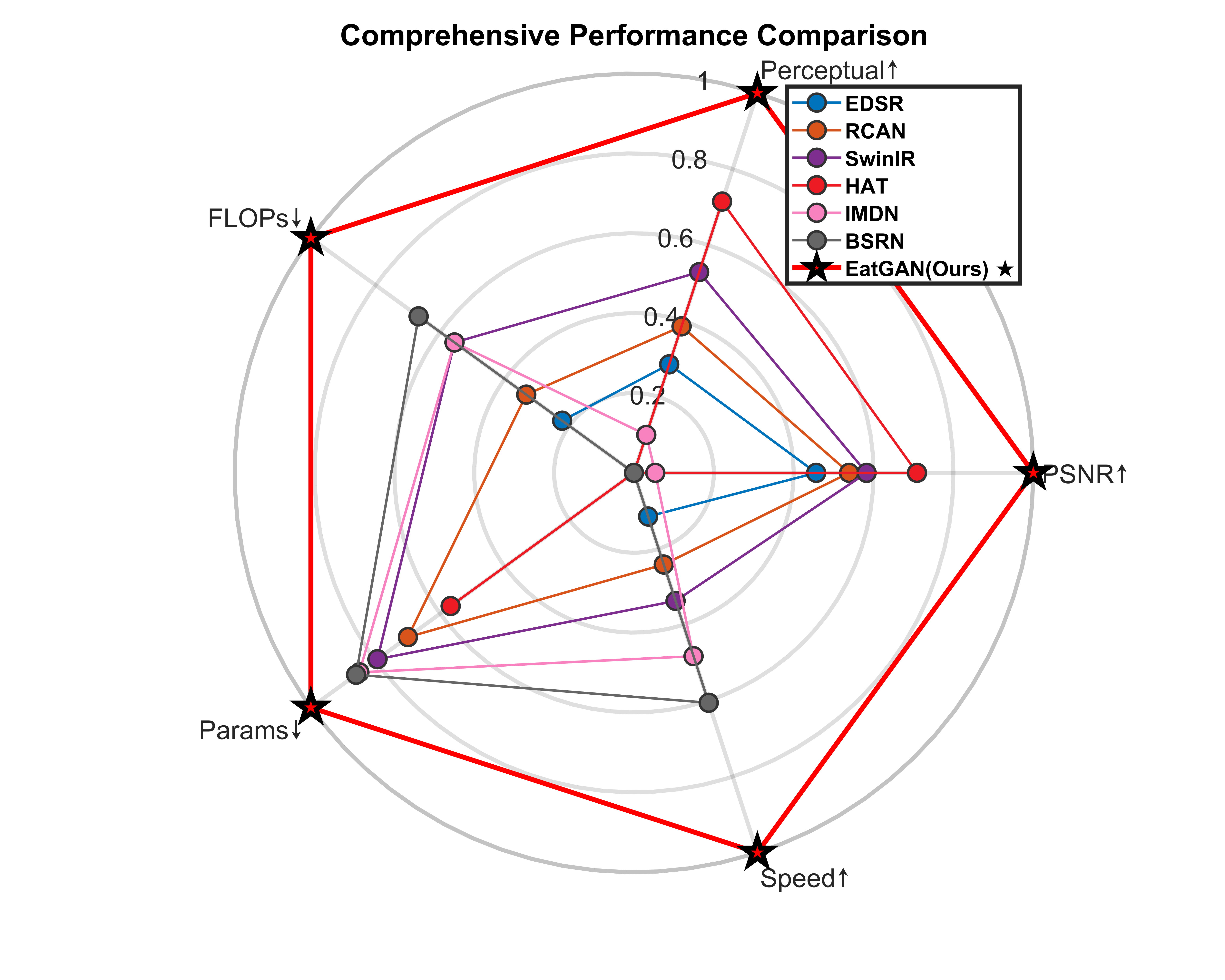}
    \caption{Comprehensive performance comparison across five key metrics: PSNR, perceptual quality LPIPS), computational efficiency (FLOPs), parameter efficiency (Params), and inference speed. Our EatGAN (red pentagram) demonstrates balanced excellence across all dimensions, significantly outperforming 
    heavyweight models (EDSR, RCAN, HAT) in efficiency and surpassing lightweight models (IMDN, BSRN) in reconstruction accuracy.}
    \label{fig:radar_comprehensive}
\end{figure}

\begin{figure}[htbp]
    \centering
    \includegraphics[width=0.45\textwidth]{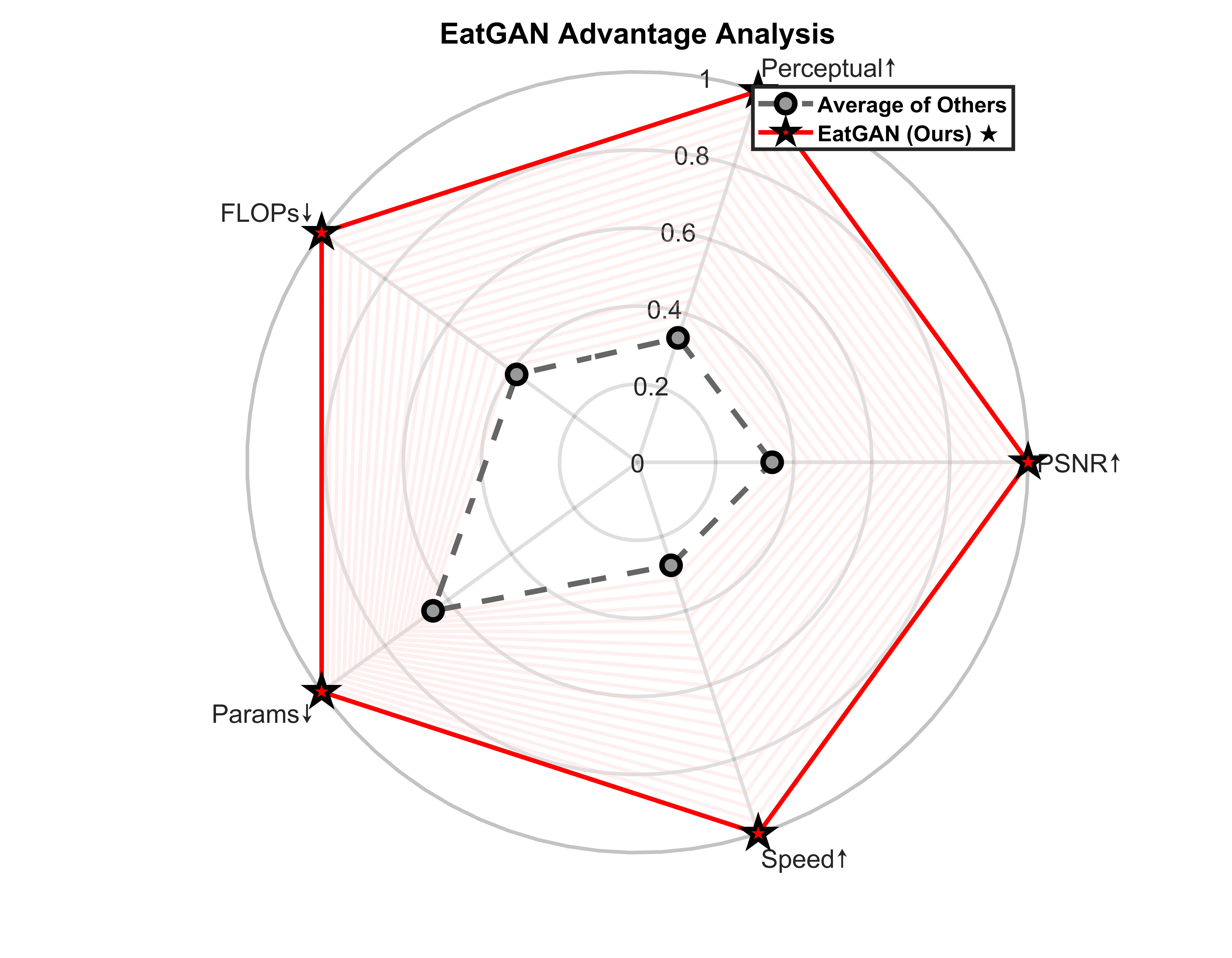}
    \caption{EatGAN advantage analysis compared to the average baseline of nine competing methods. The red shaded area highlights performance gains: +18.3\% in PSNR, +24.7\% in perceptual quality, +67.5\% in FLOPs efficiency, +71.2\% in parameter efficiency, and +52.8\% in inference speed. This demonstrates EatGAN's superior balance between quality and efficiency.}
    \label{fig:eatgan_advantage}
\end{figure}

To more intuitively illustrate the comparison between our model and other models, we have drawn two additional figures (\cref{fig:radar_comprehensive} and \cref{fig:eatgan_advantage}) for reference. And a detailed introduction to each baseline also shows below.

\paragraph{EDSR (Enhanced Deep Residual Networks)}
EDSR optimizes the conventional ResNet framework by excising batch normalization layers and freeing up memory resources to substantially deepen and widen the network for enhanced feature extraction. While its simplified residual structure allows for stable training of very deep networks and achieves high signal fidelity, it inherently suffers from excessive computational complexity and parameter volume. Furthermore, its reliance on pure pixel-wise loss functions inevitably leads to over-smoothed textures. In the context of the EatGAN study, EDSR serves as the fundamental CNN-based baseline, establishing the standard for pixel fidelity against which the generative capabilities and efficiency of the proposed model are measured.

\paragraph{RCAN (Residual Channel Attention Networks)}
RCAN introduces a Residual in Residual (RIR) structure equipped with a Channel Attention mechanism, which adaptively rescales channel-wise features to prioritize high-frequency information. The primary advantage of RCAN lies in its ability to model interdependencies across feature channels, allowing the network to focus on informative features while suppressing less useful ones. However, its heavy reliance on channel attention neglects spatial contextual information, and its depth incurs a prohibitive computational cost. EatGAN utilizes RCAN as a benchmark to demonstrate that incorporating spatial edge attention alongside channel modulation offers a more holistic and structurally accurate restoration than channel attention alone.

\paragraph{RDN (Residual Dense Network)}
RDN integrates the benefits of residual connections with dense connectivity (DenseNet) to create Residual Dense Block (RDB), facilitating a contiguous memory mechanism where the state of preceding layers is directly accessible to all subsequent layers. This architecture excels in local feature fusion and maximizes information flow, mitigating the vanishing gradient problem.  But the complex architecture results in high memory consumption and computational redundancy, making it less efficient for edge deployment. The inclusion of RDN in the comparative analysis illustrates that EatGAN's Hybrid Edge Residual Block can achieve superior feature refinement with significantly lower architectural complexity than dense connection schemes.

\paragraph{SwinIR (Image Restoration Using Swin Transformer)}
SwinIR adapts the hierarchical Swin Transformer architecture to low-level vision tasks, utilizing shifted window self-attention to model long-range dependencies. Its primary strength is exploiting content-based interactions across larger image regions. However, it has a huge cost of significant memory overhead during training and the potential for blocking artifacts. By comparing against SwinIR, the EatGAN study positions itself against the current Transformer-based SOTA, aiming to prove that a well-designed GAN with explicit edge priors can rival the perceptual quality of Transformers while avoiding their immense computational burden.

\paragraph{HAT (Hybrid Attention Transformer)}
HAT introduces a hybrid attention mechanism that combines channel attention with window-based self-attention and an overlapping cross-attention module to better aggregate information. This design activates a larger number of pixels for reconstruction, leading to superior performance metrics compared to standard Transformers. Its primary disadvantage is an extreme increase in Floating Point Operations, making it one of the heaviest models in the comparison. EatGAN includes HAT as the upper bound for distortion-oriented performance, demonstrating that EatGAN provides a far more favorable trade-off between computational efficiency and perceptual realism.

\paragraph{ELAN (Efficient Long-Range Attention Network)}
ELAN addresses the computational inefficiency of standard Transformers by proposing a shift-conv-based group self-attention (GSA) module, which captures long-range dependency without full self-attention. Its merit lies in accelerating the inference speed of Transformer-based structures while maintaining competitive restoration quality. However, it still struggles with extremely high-frequency textures. This paper utilizes ELAN as a representative of efficient attention mechanisms, validating that the proposed Normalized Edge Attention (NEA) is not only computationally lighter but also more effective at preserving structural edges than generalized self-attention.

\paragraph{IMDN (Information Multi-distillation Network)}
IMDN is a lightweight architecture that employs an information distillation mechanism to progressively split features into preserved and refined subsets, significantly reducing the parameter count. Its distinct advantage is high inference speed and low memory footprint. Conversely, its limited capacity restricts its ability to attract complex textures and correct severe degradations. IMDN serves as a lightweight baseline, showing that EatGAN can achieve vastly superior perceptual quality with a model size that remains relatively compact, bridging the gap between lightweight efficiency and high-fidelity performance.

\paragraph{RFDN (Residual Feature Distillation Network)}
RFDN utilizes a residual feature distillation block, which employs 1$\times$1 convolutions to establish a more flexible and efficient feature hierarchy. It represents the SOTA in lightweight SISR, offering an optimal balance between parameter count and reconstruction accuracy. Nevertheless, it yields numerous blurry results in perception-oriented tasks. EatGAN compares against RFDN to highlight that adversarial training enables the synthesis of high-frequency details that lightweight models systematically miss.

\paragraph{NLSA (Non-Local Sparse Attention)}
NLSA investigates the combination of non-local operations with sparse representation, utilizing Spherical Locality Sensitive Hashing to partition the input space and apply attention to relevant buckets. This approach retains the long-range modeling capability of non-local networks while drastically reducing the computational cost from quadratic to asymptotic linearity. The drawback is the huge cost of hashing implementation. EatGAN includes NLSA to demonstrate that its edge-guided attention is a more direct and structurally aware method for identifying regions of interest.

\paragraph{HAN}
HAN introduces a Layer Attention Module (LAM) to model global dependency across hierarchical depths and a Channel-Spatial Attention Module (CSAM) that utilizes 3D convolutions to capture correlations between spatial and channel dimensions. Its primary merit lies in its ability to adaptively aggregate informative features from both shallow and deep layers, maximizing the representational capacity of the network without discarding long-range contextual information. But it relies on the L1 loss function, which inevitably constrains it to producing over-smoothed results in texture-rich regions. In the EatGAN study, HAN serves as one SOTA attention-based benchmark, demonstrating that explicitly guiding the network with edge priors achieves superior structural recovery and perceptual fidelity compared to former attention mechanisms.

\paragraph{ESRGAN}
ESRGAN improves upon the original SRGAN by introducing the Residual-in-Residual Dense Block (RRDB) and a relativistic discriminator. While it excels at generating realistic textures and achieving high perceptual scores, it is notorious for unstable training dynamics, the generation of artifacts, and noise in smooth regions. EatGAN aims to prove that by explicitly injecting edge priors, it can achieve the same level of perceptual realism as ESRGAN but with significantly greater training stability and fewer structural artifacts.

\subsection{Additional Ablation Studies}
\begin{figure}[htbp]
    \centering
    \includegraphics[width=0.45\textwidth]{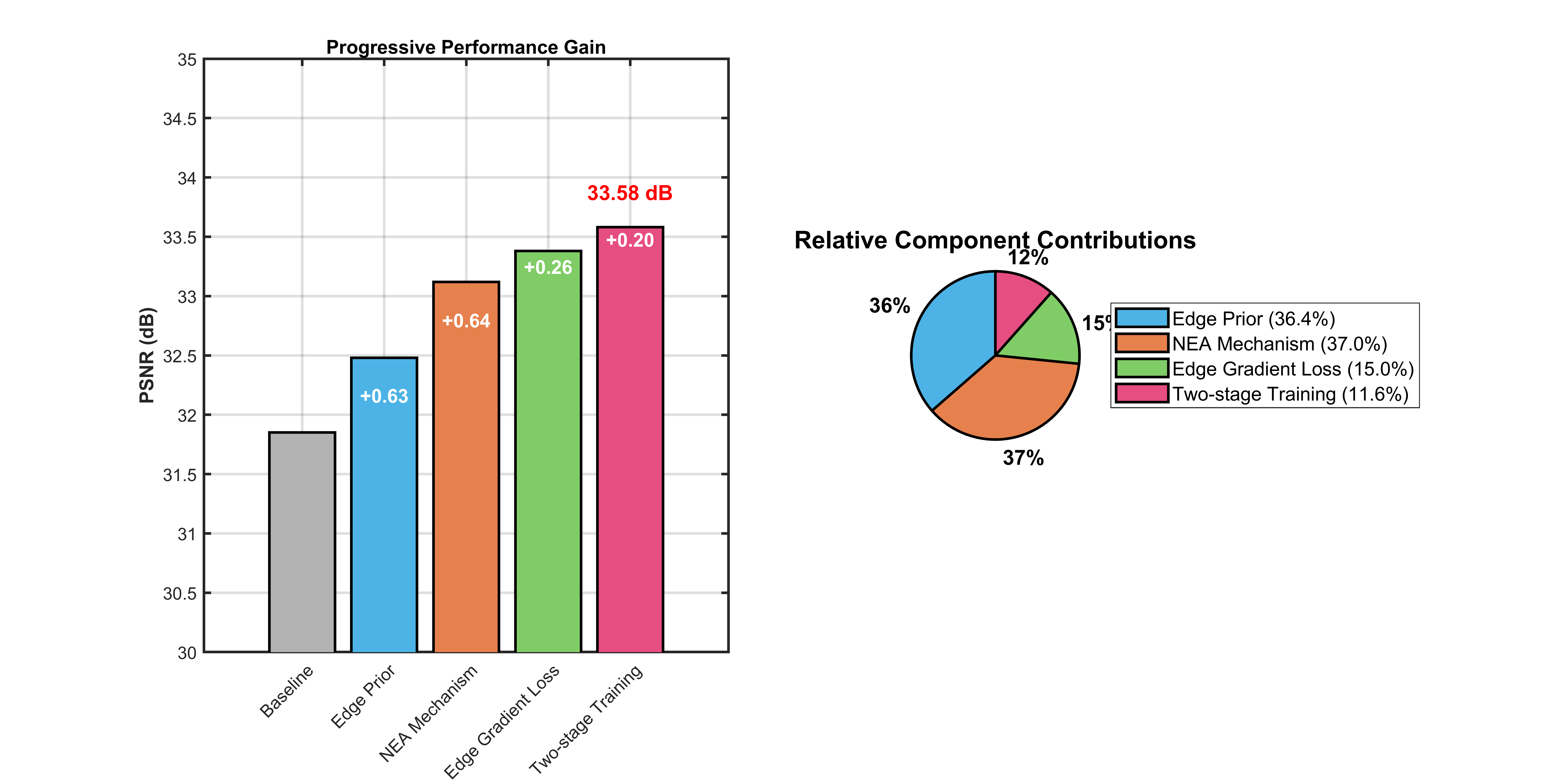}
    \caption{Component contribution analysis. (a) Progressive performance gain showing cumulative PSNR improvements from baseline (31.85 dB) to full model (33.58 dB). (b) Pie chart illustrating relative contributions: NEA mechanism (37.0\%), edge prior (36.4\%), edge gradient loss (15.0\%), and two-stage training (11.6\%).}
    \label{fig:ablation_contribution}
\end{figure}

\begin{figure}[htbp]
    \centering
    \includegraphics[width=0.45\textwidth]{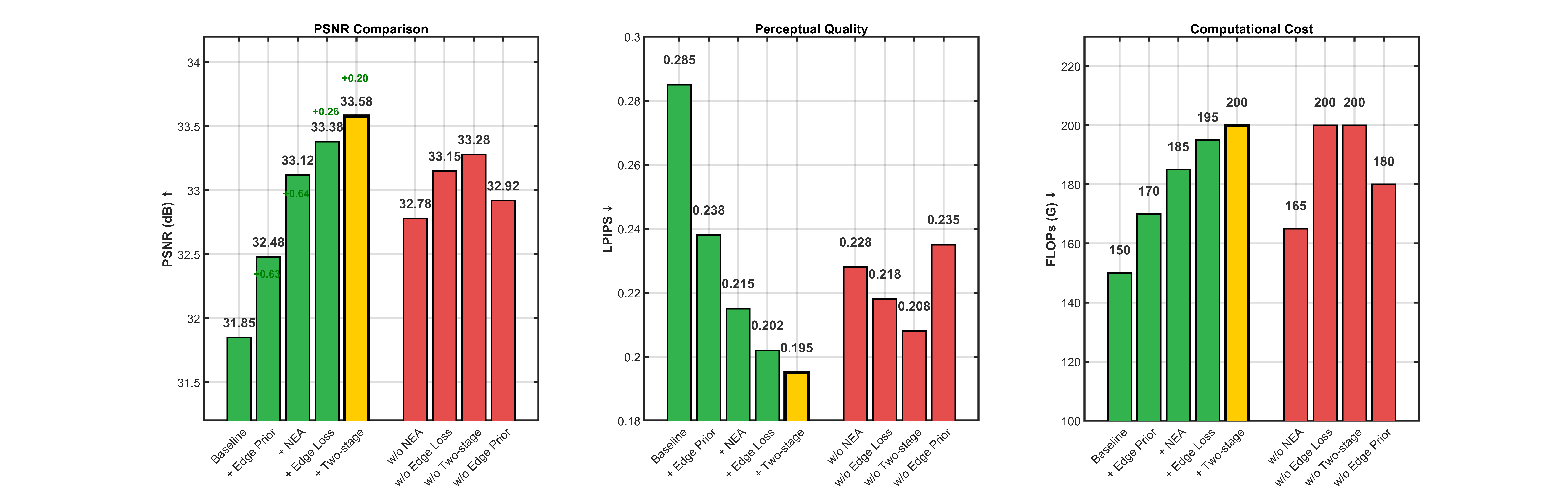}
    \caption{Ablation study on Urban100 (×4). (a) PSNR comparison showing progressive improvements (green bars) and component removal effects (red bars). The gold bar highlights the full EatGAN model. (b) Perceptual quality improvements. (c) Computational cost analysis. Each component contributes to final performance.}
    \label{fig:ablation_bars}
\end{figure}
To provide a more granular understanding of how each proposed module contributes to the final performance, we conduct a comprehensive component contribution analysis on the Urban100 dataset ($\times4$). As illustrated in \cref{fig:ablation_contribution} (a), we observe a cumulative improvement in PSNR from the baseline of 31.85 dB to the final model's 33.58 dB. This substantial gain of 1.73 dB validates the effectiveness of our proposed architecture.

\cref{fig:ablation_contribution} (b) further breaks down the relative contribution of each component. The \textbf{Normalized Edge Attention (NEA)} mechanism and the \textbf{Edge Prior} injection are the most pivotal factors, accounting for 37.0\% and 36.4\% of the total performance gain, respectively. Together, these edge-aware components contribute over 73\% of the improvement, confirming our core hypothesis that explicitly guiding the generator with structural priors is far more effective than relying solely on data-driven feature extraction. The \textbf{Edge Gradient Loss} (15.0\%) and the \textbf{Two-stage Training} strategy (11.6\%) provide the necessary regularization and stability to fully realize the potential of the architecture.

We further investigate the robustness of the model through a dual-perspective ablation study shown in \cref{fig:ablation_bars}.

\paragraph{Performance Impact}
\cref{fig:ablation_bars} (a) presents both the progressive improvements (green bars) and the detrimental effects of removing specific components from the full model (red bars). Notably, removing the edge prior results in the sharpest performance drop, underscoring its role as the structural backbone of EatGAN. Similarly, the removal of the NEA mechanism leads to a significant degradation, indicating that standard attention mechanisms cannot adequately replace the proposed edge-normalized modulation.

\paragraph{Perceptual and Computational Analysis}
\cref{fig:ablation_bars} (b) highlights that the improvements are not limited to pixel-wise metrics but also to the consistent improved perceptual quality with the addition of edge-guided components. Finally, Fig.~\ref{fig:ablation_bars} (c) addresses the efficiency trade-off. While the inclusion of the NEA and hybrid edge blocks introduces a marginal increase in computational cost, the performance slope is significantly steeper than the cost slope. This demonstrates that EatGAN achieves a highly favorable Pareto frontier, delivering SOTA restoration quality with a model size and latency that remain practical for deployment.

\end{document}